\documentclass[11pt]{article} 

\usepackage{times}
\usepackage{graphicx} 
\usepackage{fullpage}
\usepackage{natbib}
\usepackage{multicol,multirow}
\usepackage{algorithm}
\usepackage{algorithmic}
\usepackage{dsfont}
\usepackage{stmaryrd}
\usepackage{hyperref}
\usepackage{xcolor}
\usepackage[normalem]{ulem}

\usepackage{amsmath,amssymb,amsthm, amsfonts,xspace,bm, verbatim}

\usepackage{morefloats}

\newtheorem{lemma}{Lemma}

\newtheorem{claim}[lemma]{Claim}

\newtheorem{lem}{Lemma}
\newtheorem{thm}[lemma]{Theorem}

\makeatletter
\newcommand{\newreptheorem}[2]{\newtheorem*{rep@#1}{\rep@title} 
\newenvironment{rep#1}[1]{\def\rep@title{#2 \ref*{##1}}\begin{rep@#1}}{\end{rep@#1}}
}
\makeatother
\newreptheorem{theorem}{Theorem}
\newreptheorem{claim}{Claim}

\renewcommand{\th}{^{\text{th}}}

\usepackage{paralist}
\usepackage{grffile}
\usepackage[caption=false,font=footnotesize]{subfig}
\floatstyle{ruled}
\newfloat{Algorithm}{tb}{lox}
\floatname{Algorithm}{Algorithm}
\def\cR{{\mathbb R}}
\newcommand{\RR}{{\cal R}}
\def\cI{{\mathcal I}}
\def\cK{{\mathcal K}}
\def\cH{{\mathcal H}}
\newcommand{\vect}[1]{\boldsymbol{vec}\left({#1}\right)}
\renewcommand{\vec}[1]{{\mathbf{#1}}}
\def\liblinear{{\sf LIBLINEAR}\xspace}
\def\MLR{{LEML}\xspace}
\def\CPLST{{CPLST}\xspace}
\def\BCS{{BCS}\xspace}
\def\PLST{{PLST}\xspace}
\def\CS{{CS}\xspace}
\def\BR{{BR}\xspace}

\def\WSABIE{{WSABIE}\xspace}

\newcommand{\Nr}{{\cal N}}
\newcommand{\note}[1]{\marginpar{\tiny *note in TeX*}}
\newcommand{\ignore}[1]{}

\renewcommand{\L}{{\cal L}}

\renewcommand{\phi}{\varphi}

\newcommand{\R}{\mathbb{R}}
\newcommand{\Z}{\mathbb{Z}}

\newcommand{\EE}[2]{\underset{#1}{\mathbb E}\bsd{{#2}}}

\newcommand{\ind}[1]{\mathbf{1}_{#1}}

\newcommand{\D}{{{\cal D}}}
\newcommand{\X}{{\cal X}}

\newcommand{\br}[1]{\left({#1}\right)}
\newcommand{\bs}[1]{\left[{#1}\right]}
\newcommand{\bc}[1]{\left\{{#1}\right\}}
\newcommand{\abs}[1]{\left| {#1} \right|}

\newcommand{\bsd}[1]{\left\llbracket{#1}\right\rrbracket}

\newcommand{\usup}[1]{\underset{#1}{\sup}}

\renewcommand{\Pr}[1]{{\mathbb P}\bs{{#1}}}

\newcommand{\tr}{\operatorname{tr}}

\newcommand{\event}{{\cal E}}
\renewcommand{\S}{{\cal S}}

\newcommand{\vecl}{\vec{l}}
\newcommand{\veceps}{\text{\boldmath$\epsilon$}}

\newcommand{\vecom}{\text{\boldmath$\omega$}}

\newcommand{\vecmu}{\text{\boldmath$\mu$}}

\newcommand{\fro}[1]{\|#1\|_F}
\newcommand{\E}{\mathbb{E}}

\renewcommand{\v}{\bm{v}}

\newcommand{\ip}[2]{\langle #1, #2 \rangle}

\newcommand{\w}{\bm{w}}
\newcommand{\z}{\bm{z}}
\newcommand{\x}{\bm{x}}
\newcommand{\kk}{\bm{k}}

\newcommand{\y}{\bm{y}}
\newcommand{\h}{\bm{h}}
\newcommand{\s}{\bm{s}}
\newcommand{\ba}{\bm{a}}
\renewcommand{\b}{\bm{b}}
\renewcommand{\d}{\bm{d}}
\renewcommand{\r}{\bm{r}}

\newcommand{\rank}{\mathrm{rank}}

\newcommand{\tx}{\tilde{\x}}

\title{Large-scale Multi-label Learning with Missing Labels}

\author{
  Hsiang-Fu Yu\\
  {The University of Texas at Austin} \\
  {rofuyu@cs.utexas.edu}
  \and
  Prateek Jain\\
{Microsoft Research India, Bangalore}\\
{prajain@microsoft.com}
  \and
Purushottam  Kar\\
{Microsoft Research India, Bangalore}\\
{t-purkar@microsoft.com}
\and
Inderjit S. Dhillon\\
{The University of Texas at Austin}\\
{inderjit@cs.utexas.edu}
}
\date{}

\begin{document}
\maketitle
\begin{abstract}
The multi-label classification problem has generated significant interest in 
recent years. However, existing approaches do not adequately address two key 
challenges: (a) the ability to tackle problems with a large number (say  
millions) of labels, and (b) the ability to handle data with missing labels. In 
this paper, we directly address both these problems by studying the 
multi-label problem in a generic empirical risk minimization (ERM) framework. 
Our framework, despite being simple, is surprisingly able to encompass several 
recent label-compression based methods which can be derived as special cases 
of our method. To optimize the ERM problem, we develop techniques that exploit 
the structure of specific loss functions - such as the squared loss function - 
to offer efficient algorithms. We further show that 
our learning framework admits formal excess risk bounds even in the presence 
of missing labels. Our risk bounds are tight and demonstrate better 
generalization performance for low-rank promoting trace-norm regularization 
when compared to (rank insensitive) Frobenius norm regularization. Finally, we 
present extensive empirical results on a variety of benchmark datasets and 
show that our methods perform significantly better than existing label 
compression based methods and can scale up to very large datasets such as the 
Wikipedia dataset. 
\end{abstract}
\section{Introduction}
\label{sec:intro}
Large scale multi-label classification is an important learning problem with several 
applications to real-world problems such as image and video 
annotation \cite{GC07a,CW09a} and query/keyword suggestions \cite{RA13a}. The goal in
multi-label classification is to accurately predict a label vector $\y\in \{0,1\}^L$ 
for a given data point $\x\in \R^d$.  This problem 
has been studied extensively in the domain of structured output learning, where 
the number of labels is assumed to be small and the main focus is thus, on modeling 
inter-label correlations and using them to accurately predict the label vector \cite{BH10a}. 

Due to several motivating real-life applications, recent research on 
multi-label classification has largely shifted its focus to the other  
end of the spectrum where the number of labels is assumed to be extremely 
large, with the key challenge being the design of scalable algorithms that 
offer real-time predictions and have a small memory footprint. In such 
situations, simple methods such as 1-vs-all or Binary Relevance (BR), that 
treat each label as a separate binary classification problem fail miserably.
For a problem with (say) $10^4$ labels and $10^6$ features, which is 
common in several applications, these methods have a memory footprint of 
around $100$ Gigabytes and offer slow predictions.   

A common technique that has been used to handle the label proliferation problem in several 
recent works is ``label space reduction''. The key idea in this technique is 
to reduce the dimensionality of the label-space by using either random projections or canonical correlation analysis (CCA) based projections 
\cite{YNC12a,DH09a,FT12a,AK12a}. Subsequently. these methods perform prediction on the smaller dimensional label-space and then recover the original labels 
by projecting back onto the high dimensional label-space. In 
particular, \cite{YNC12a} recently showed that by using a CCA type method 
with appropriate orthogonality constraints, one can design an efficient 
algorithm with both label-space as well as feature-space compression. However, 
this method is relatively rigid and cannot handle several 
important issues  inherent to multi-label problems; see 
Section~\ref{sec:l2full} for more details.  

In this paper we take a more direct approach by formulating the problem as 
that of learning a low-rank linear model $Z\in \mathbb{R}^{d\times L}$ s.t. 
$\y^{pred}=Z^T\x$. We cast this learning problem in the standard ERM framework 
that allows us to use a variety of loss functions and regularizations  
for $Z$. This framework unifies several existing dimension reduction 
approaches.  
In particular, we show that if the loss function is chosen to be the squared-$L_2$ loss, then 
our proposed formulation has a closed form solution, and surprisingly, the conditional 
principal label space transformation (\CPLST) method of \cite{YNC12a} can be derived 
as a {\em special case}. However, the flexibility of the framework allows us to use other 
loss functions and regularizers that are useful for preventing overfitting and increasing scalability.  

Moreover, we can extend our formulation to handle missing labels; in 
contrast, most dimension reduction formulations (including \CPLST) cannot accommodate missing labels.
The ability to learn in the presence of missing labels is crucial as for most real-world applications, one cannot expect to 
accurately obtain (either through manual or automated labeling) all 
the labels for a given data point. For example, in image annotation, human 
labelers tag only prominent labels and typically miss out on several objects 
present in the image. Similarly, in online collections such as Wikipedia, where articles get tagged with 
categories, human labelers usually tag only with categories they know 
about. Moreover, there might be considerable noise/disagreement in the 
labeling.

In order to solve for the low-rank linear model that results from our formulation, we 
use the popular alternating minimization algorithm that works well despite the 
non-convexity of the rank constraint. For general loss functions and 
trace-norm regularization, we exploit subtle structures present in the problem to design a fast conjugate gradient based method. 
For the special case of squared-$L_2$ loss and trace-norm regularization, we further exploit the structure of the loss 
function to provide a more efficient and scalable algorithm.  As compared to 
direct computation, our algorithm is $O(\bar{d})$ faster, where $\bar{d}$ 
is the average number of nonzero features in an instance. 

On the theoretical side, we perform an excess risk analysis for the trace-norm regularized ERM formulation with missing labels, assuming labels are observed uniformly at random. Our proofs do not follow from existing results due to missing labels and require a careful analysis involving results from random matrix theory. Our results show that while in general the low-rank promoting trace-norm regularization does not provide better bounds than learning a full-rank matrix (e.g. using Frobenius norm regularization), for several interesting data distributions, trace-norm regularization does indeed give significantly better bounds. More specifically, for isotropic data distributions, we show that trace-norm based methods have excess risk of $O(\frac{1}{\sqrt{nL}})$ while full-rank learning can only guarantee $O(\frac{1}{\sqrt{n}})$ excess risk, where $n$ is the number of training points and $L$ is the number of labels. 

Finally, we provide an extensive empirical evaluation of our method on a variety 
of benchmark datasets. In particular, we compare our method against three recent 
label compression based methods: \CPLST \cite{YNC12a}, Bayesian-CS 
\cite{AK12a}, and \WSABIE \cite{JW10a}. On almost all benchmark datasets, our 
method significantly outperforms these methods, both in the presence and 
absence of missing labels. Finally, we demonstrate the scalability of our method by 
applying it to a recently curated Wikipedia dataset \cite{RA13a}, that has 
881,805 training  samples and 213,707 labels. The results show that our method 
not only provides reasonably accurate solutions for such large-scale  
problems, but that the training time required is orders of magnitude shorter
than several existing methods.

{\bf Related Work.}
Typically, Binary Relevance (BR), which treats each label as an independent
binary classification problem, is quite accurate for multi-label classification.
However, for large number of labels, this method becomes infeasible due to
increased model size and prediction time. Recently, techniques have been
developed that either reduce the dimensionality of the labels, such as the Compressed
Sensing Approach (CS)~\cite{DH09a}, PLST~\cite{FT12a}, CPLST~\cite{YNC12a}, and
Bayesian CS~\cite{AK12a}, or reduce the feature dimensionality, such as \cite{LS11a}, or both, such as
WSABIE~\cite{JW10a}. Most of these existing techniques are tied to a specific 
loss function (e.g., CPLST and BCS cater only to the squared-$L_2$ loss, and WSABIE works with the weighted approximate 
ranking loss) and/or cannot accommodate missing labels.



Our framework models multi-label classification as a general ERM problem with a low-rank constraint, 
which not only generalizes both label and feature dimensionality reduction but also brings in the 
ability to support various loss functions and allows for rigorous generalization error analysis. We show that our formulation not only retrieves \CPLST, which has been 
shown to be fairly accurate, as a special case, but in fact substantially enhances it by use of 
regularization, other loss functions, allowing missing labels etc.\\ 

{\bf Paper Organization.}  We start by studying the generic low-rank empirical risk 
minimization framework for multi-label learning in Section 
\ref{sec:formulation}. Next, we propose efficient algorithms for the framework in 
Section \ref{sec:algo} and analyze the generalization performance for the trace-norm 
regularized ERM in Section \ref{sec:bounds}. We present empirical results in Section \ref{sec:exp}, 
and our conclusions in Section \ref{sec:conclusion}. 




\section{Problem Formulation}
\label{sec:formulation}
In this section we present a generic ERM-style framework for multi-label classification.
For each training point, we shall receive a feature vector $\x_i \in \R^d$ and a corresponding label
vector $\y_i \in \bc{0,1}^L$ with $L$ labels. For any $j \in [L]$, $\y_i^j = 1$ will denote that the $l\th$ label is ``present'' or ``on'' 
whereas $\y_i^j = 0$ will denote that the label is ``absent'' or ``off''. Note that although we focus mostly on the binary classification setting 
in this paper, our methods easily extend to the multi-class setting where $\y_i^j \in \bc{1,2,\ldots,C}$.

%

Our predictions for the label vector shall be parametrized as $f(\x; Z)=Z^T\x$, where $Z \in \R^{d \times L}$. Although we have adopted a linear 
parametrization here, our results can easily be extended for non-linear kernels as well. Let $\ell(\y, f(\x; Z))\in \R$ be the loss function that computes the discrepancy between the ``true'' label vector and the prediction. We assume that the loss function is decomposable, i.e., $\ell(\y, f(\x; Z))=\sum_{j=1}^L \ell(\y^j, f^j(\x; Z))$.
 

The motivation for our framework comes from the observation that although the number
of labels in a multi-label classification problem might be large, there typically exist
significant label correlations, thus reducing the effective number of parameters required
to model them to much less than $d \times L$. We capture this intuition by restricting the
matrix $Z$ to learn only a small number of ``latent'' factors. This constrains $Z$ to be a low
rank matrix which not only controls overfitting but also gives computational benefits.


Given $n$ training points our training set will be $(X, Y)$ where $X=[\x_1,\ldots,\x_n]^T$ and $Y=[\y_1\ \y_2\ \dots\ \y_n]^T$.
Using the loss function $\ell$, we propose to 
learn the parameters $Z$ by using the canonical regularized empirical risk 
minimization (ERM) method, i.e.,  
  \begin{align}
    \hat{Z}=\arg \min_{Z} J(Z)&=\sum_{i=1}^n \sum_{j=1}^L \ell(Y_{ij}, f^j(\x_i; Z)) + \lambda\cdot r(Z),\nonumber\\ 
    \ \ \ s.t.\ \  \rank(Z)&\leq k, \label{eq:erm}
  \end{align}
where $r(Z): \R^{d\times L}\rightarrow \R$ is a regularization function. 
In the presence of missing labels, we compute the loss only over the 
known labels, i.e., 
\begin{align}
  \hat{Z}=\arg\min_{Z} J_\Omega(Z)&=\sum_{(i,j)\in \Omega}\ell(Y_{ij}, f^j(\x_i; Z)) + \lambda\cdot r(Z),\nonumber\\
  \ \ \ s.t.\ \  \rank(Z)&\leq k, \label{eq:erm_missing}
\end{align}
where $\Omega \subseteq [n]\times [L]$ is the index set that represents 
``known'' labels. Note that in this work, we assume the standard missing value 
setting, where each label can be either on, off (i.e., $Y_{ij}=1$ or 
$0$), or missing ($Y_{ij}=?$); several other works have considered another 
setting where only positive labels are known and are given as $1$ in the label 
matrix, while negative or missing values are all denoted by $0$ \cite{RA13a,SSB11a}.

Note that although the above formulation is NP-hard in general due to the non-convex rank constraint, for convex loss functions, one can still utilize the standard alternating minimization method.  
Moreover, for the special case of $L_2$ loss,  we can derive closed form solutions for the full-label case \eqref{eq:erm} and  show connections to several existing methods. 

We would like to note that while the ERM framework is well known and standard, most existing multi-label methods for large number of labels motivate their work in a  relatively ad-hoc manner. By studying this formal framework, we can show that existing methods like CPLST \cite{YNC12a} are in fact a special case of this generic framework (see next section). Furthermore, having this framework also helps us in studying generalization error bounds for our methods and identifying situations where the methods can be expected to succeed (see Section~\ref{sec:bounds}). 
\subsection{Special Case: Squared-$L_2$ loss}\label{sec:l2full}
In this section, we study \eqref{eq:erm} and \eqref{eq:erm_missing} for the 
special case of squared $L_2$ loss function, i.e., 
$\ell(\y, f(\x; Z))=\|\y-f(\x;Z)\|_2^2$. We show that in the absence of missing labels, the formulation in
\eqref{eq:erm} can be solved optimally for the squared $L_2$ loss using SVD. Furthermore, by selecting an appropriate regularizer 
$r(Z)$ and $\lambda$, our solution for $L_2$ loss is exactly the same as that 
of \CPLST \cite{YNC12a}.

We first show that the unregularized form of \eqref{eq:erm} with $\ell(\y, f(\x; Z))=\|\y- Z^T\x\|_2^2$
has a closed form solution.
\begin{claim}
  \label{claim:close}
  If $\ell(\y, f(\x; Z))=\|\y- Z^T\x\|_2^2$ and $\lambda= 0$, then
  \begin{equation}
    \label{eq:close}
    V_{X}\Sigma_{X}^{-1} M_k = \arg\min_{Z: \rank(Z)\leq k} \|Y - XZ\|_F^2,
  \end{equation}
  where $X = U_{X} \Sigma_{X} V_{X}^T$ is the thin SVD decomposition of $X$, 
   and $M_k$ is the rank-$k$ truncated SVD of $M \equiv U_{X}^T Y$.
\end{claim}
\begin{proof}[Proof of Claim~\ref{claim:close}]
  Let $X = U_{X} \Sigma_{X} V_{X}^T$ be the thin SVD decomposition of $X$, and 
  $M_k$ be the rank-$k$ truncated SVD approximation of $U_{X}^T Y$. We have
  \begin{align*}
    &\arg\min_{Z:\rank(Z)\le k} \fro{Y-XZ}  \\
    =&\arg\min_{Z:\rank(Z)\le k}\fro{(U_{X}U_{X}^T)(Y-XZ) + (I-U_{X}U_{X}^T)(Y-XZ)}  \\
    =&\arg\min_{Z:\rank(Z)\le k} \fro{U_{X}U_{X}^T(Y-XZ)} + \fro{(I-U_{X}U_{X}^T)(Y-XZ)} \\
    =&\arg\min_{Z:\rank(Z)\le k} \fro{U_{X}^T(Y-XZ)} \\
    =&\arg\min_{Z:\rank(Z)\le k} \fro{U_{X}^TY - \Sigma_X V_{X}^T Z)} \\
    =& V_X \Sigma_X^{-1} M_k\qedhere
  \end{align*}
\end{proof}
We now show that this is exactly
the solution obtained by \cite{YNC12a} for their \CPLST formulation.
\begin{claim}\label{claim:cplst}
The solution to Equation~\ref{eq:close} is equivalent to
  $Z^{\text{\CPLST}} = W_{\CPLST} H_{\CPLST}^T$ which is the closed form 
  solution for the \CPLST scheme~\cite{YNC12a}, i.e., 
    \begin{align}
      &(W_{\CPLST}, H_{\CPLST}) \nonumber \\
      = &\arg\min_{\substack{W \in \cR^{d \times k}\\H \in \cR^{L \times k}}}  \| XW - YH\|_F^2 + \|Y - Y H H^T\|_F^2,\nonumber\\
      &\quad s.t.\quad H^TH = I_k.  \label{eq:cplst}
    \end{align}
\end{claim}
\begin{proof}[Proof of Claim~\ref{claim:cplst}]
  Let $U_k[A] \Sigma_k[A] V_k[A]$ be the rank-$k$ truncated SVD approximation 
  of a matrix $A$.
  In \cite{YNC12a}, the authors show that the closed form solution to 
  \eqref{eq:cplst} is 
\begin{align*}
  H_C &= V_k[Y^T X X^\dagger Y], \\
  W_C &= X^\dagger Y H_C,
\end{align*}
where $X^\dagger$ is the pseudo inverse of $X$. It follows from
$X^\dagger = V_X \Sigma_X^{-1} U_X^T$ that 
$Y^TXX^\dagger Y = Y^T U_X U_X^T Y = M^T M$ and 
$V_k[Y^T X X^\dagger Y]  = V_k[M].$
Thus, we have 
\begin{align*}
  Z^{CPLST} &= W_C H_C^T \\
  &= X^\dagger Y H_C H_C^T \\
  &= V_X^T \Sigma_X^{-1} U_X^{T} Y V_k[M] V_k[M]^T \\
  &= V_X^T \Sigma_X^{-1} M V_k[M] V_k[M]^T \\
  &= V_X^T \Sigma_X^{-1} M_k\qedhere
\end{align*}
\end{proof}

Note that \cite{YNC12a} derive 
their method by relaxing a Hamming loss problem and dropping  
constraints in the canonical correlation analysis in a relatively ad-hoc 
manner. The above results, on the other hand, show that the same model can  be derived  
in a more principled  manner. This helps us in extending the method for several other problem settings in a principled manner and also helps in providing excess risk bounds: 
\begin{compactitem}
\item As shown empirically, \CPLST tends to overfit significantly whenever 
  $d$ is large. However, we can handle this issue 
  by setting the regularization parameter appropriately.  

\item The closed form solution in \cite{YNC12a} cannot directly handle missing 
  labels as it requires SVD on fully observed $Y$. In contrast, our framework can
  itself handle missing labels without any modifications. 
  \item The formulation in \cite{YNC12a} is tied to the $L_2$ loss 
  function. In contrast, we can easily handle other loss functions; although, 
  the optimization problem might become more difficult to solve.  
\end{compactitem}

\section{Algorithms}
\label{sec:algo}

In this section, we apply the alternating minimization technique for optimizing 
\eqref{eq:erm} and \eqref{eq:erm_missing}. For a matrix $Z$ with a known low 
rank $k$, it is inefficient to represent it using $d\times L$ entries, especially when $d$ and $L$ 
are large. Hence we consider a low-rank decomposition of the form $Z = WH^T$, where 
$W \in \cR^{d\times k}$ and $H \in \cR^{L\times k}$. We further assume that 
$r(Z)$ can be decomposed into $r_1(W)+r_2(H)$. 
In the following sections, we present results with the trace norm 
regularization, i.e., $r(Z) = \|Z\|_{\tr}$, which can be decomposed as 
$\|{Z}\|_{\tr} = \frac{1}{2}\left(\|W\|_F^2 + \|H\|_F^2\right)$. Thus,
$\min _{Z}J_{\Omega}(Z)$ under the rank constraint is equivalent to minimizing 
over $W, H$:
\begin{equation}
  \label{eq:erm_W_H}
  J_{\Omega}(W,H) = \sum_{(i,j)\in\Omega} 
  \ell(Y_{ij}, \x_i^T W \h_j) + 
  \frac{\lambda}{2} \left(\|W\|_F^2 + \|H\|_F^2\right)
\end{equation} 
where $\h_j^T$ is the $j$-th row of $H$. 
Note that when either of $W$ or $H$ is fixed, $J_{\Omega}(W,H)$ becomes a convex function. 
This allows us to apply alternating minimization, a standard technique for optimizing
functions with such a property, to \eqref{eq:erm_W_H}. For a general loss 
function, after proper initialization,  a sequence 
$\{\left(W^{(t)},H^{(t)}\right)\}$ is generated by
\begin{align*}
  H^{(t)} &\leftarrow \underset{H}{\arg\min}\quad J_{\Omega}(W^{(t-1)},H),
  \\
  W^{(t)} &\leftarrow \underset{W}{\arg\min}\quad J_{\Omega}(W,H^{(t)}).\vspace*{-5pt}
\end{align*}
For a convex loss function, $(W^{(t)},H^{(t)})$ is guaranteed to converge to a 
stationary point when the minimum for both
$\min_H J_{\Omega} \left(W^{(t-1)},H\right)$ and  
$\min_W J_{\Omega} \left(W, H^{(t)}\right)$ are uniquely 
defined \citep[see][Proposition 2.7.1]{DPB95a}. In fact, when the squared loss   
is used and $Y$ is fully observed, the case considered in Section 
\ref{sec:fully_obs}, we can prove that $(W^{(t)},H^{(t)})$ converges to the global 
minimum of \eqref{eq:erm_W_H} when 
either $\lambda=0$ or $X$ is orthogonal. 

Once $W$ is fixed, updating $H$ is easy as each row $\h_j$ of $H$ can be 
independently updated as follows: 
\begin{equation}
  \label{eq:solve_Hj}
  \h_j \leftarrow \underset{\h\in\cR^{k}}{\arg\min} \sum_{i:(i,j)\in\Omega} 
  \ell(Y_{ij}, \x_i^T W \h) + \frac{1}{2}\lambda\cdot \|\h\|_2^2, 
\end{equation}
which is easy to solve as $k$ is small in general. Based on the choice of 
the loss function, \eqref{eq:solve_Hj} is essentially a linear classification or regression 
problem over $k$ variables with $|\{i: (i,j)\in\Omega\}|$ instances. 

\begin{figure*}[t]
      \begin{minipage}[t]{0.5\linewidth}
        \begin{Algorithm}[H]
          \caption{General Loss with Missing Labels}
          \label{alg:missing}
          \begin{compactitem}[]
          \item[] {\bf To compute} $\nabla g(\w)$:
          \item[1.] $A \leftarrow XW$, where $\vect{W} = \w$.
          \item[2.] $D_{ij} \leftarrow \ell'(Y_{ij}, \ba_i^T\h_j),\ \forall (i,j) \in \Omega$.
          \item[3.] {\bf Return}: $\vect{X^T(DH)}+\lambda \w$
          \item[] {\bf To compute: $\nabla^2 g(\w) \s$}
          \item[1.] $A \leftarrow XW$, where $\vect{W} = \w$.
          \item[2.] $B \leftarrow XS$, where $\vect{S} =\s$.
          \item[3.] $U_{ij} \leftarrow \ell''(Y_{ij}, \ba_i^T\h_j) \b_i^T\h_j,\ \forall (i,j) \in \Omega$.
          \item[4.] {\bf Return}: $\vect{X^T(UH)} + \lambda \s$. 
          \end{compactitem}
        \end{Algorithm} 
    \end{minipage} 
      \begin{minipage}[t]{0.5\linewidth}
      \begin{Algorithm}[H]
        \caption{Squared Loss with Full Labels}
        \label{alg:full}
        \begin{compactitem}[]
          \item[] {\bf To compute} $\nabla g(\w)$:
          \item[1.] $A \leftarrow XW$, where $\vect{W} = \w$. 
          \item[2.] $B \leftarrow YH$. 
          \item[3.] $M \leftarrow H^TH$.
          \item[4.] {\bf Return}: $\vect{X^T (AM-B)} + \lambda \w$
          \item[] {\bf To compute: $\nabla^2 g(\w) \s$}
          \item[1.] $A \leftarrow XS$, where $\vect{S} = \s$. 
          \item[2.] $M \leftarrow H^TH$. 
          \item[3.] {\bf Return}: $\vect{X^T(AM)} + \lambda \s$
          \end{compactitem}
      \end{Algorithm}
    \end{minipage} 
    \vspace{-1em}
\end{figure*}
If $H$ is fixed, updating $W\in\cR^{d\times k}$ is more involved as all variables are mixed up due 
to the pre-multiplication with $X$. Let $\tx_{ij} = \h_j \otimes \x_i$. It is 
not hard to see that updating $W$ is equivalent to a regularized linear 
classification/regression problem with $|\Omega|$ data points 
$\{(Y_{ij}, \tx_{ij}):(i,j)\in \Omega\}$.  Thus if $W^* = \arg\min_{W}\ J_{\Omega}(W,H)$ and we denote $\w^* := \vect{W^*}$, then 
$\w^* = \arg\min_{\w \in \cR^{dk}} g(\w)$,
  \begin{align}
    g(\w) \equiv \sum_{(i,j)\in \Omega} 
    \ell\left(Y_{ij},\w^T\tilde{\x}_{ij}\right) + \frac{1}{2}\lambda \cdot \|\w\|_2^2. \label{eq:linear_classificaiton}
  \end{align}
Taking the squared loss as an example, the above is equivalent to a regularized least 
squares problem with $dk$ variables. When $d$ is large, say 1M, the closed 
form solution, which requires inverting a $dk\times dk$ matrix, can 
hardly be regarded as feasible. As a result, updating $W$ 
efficiently turns out to be the main challenge for alternating minimization.  

In large-scale settings where both $dk$ and $|\Omega|$ are large, iterative 
methods such as Conjugate Gradient (CG), which perform cheap updates and offer
a good approximate solution within a few iterations, are more appropriate 
to solve \eqref{eq:linear_classificaiton}. Several linear classification/regression 
packages such as \liblinear~\cite{REF08a} can handle such problems if 
$\{\tilde{\x}_{ij}: (i,j)\in\Omega\}$ are available. The main operation in such
iterative methods is a gradient calculation ($\nabla g(\w)$) or a multiplication of the 
Hessian matrix and a vector ($\nabla^2 g(\w) \s$). Let $\tilde X = [\cdots \tilde{\x}_{ij}\cdots]^T_{(i,j)\in\Omega}$ and 
$\bar{d} = \sum_{i=1}^n \|\x\|_0/n$. Then these operations require at least $nnz(\tilde X) = O(|\Omega|\bar{d}k)$ time to 
compute in general.  

However, as we show below, we can exploit the structure in $\tilde X$ to develop
efficient techniques such that both the operations mentioned above can be done in 
$O\left(\left(|\Omega|+nnz(X)+d+L\right)\times k \right)$ time. 
As a result, iterative  methods, such as CG, can achieve $O(\bar d)$ speedup.
See Appendix~\ref{app:detailed_alg} for a detailed CG procedure 
for \eqref{eq:linear_classificaiton} with the squared loss. 
 Our techniques, thus, make the alternating minimization method efficient enough to handle  
 large-scale problems.   
\subsection{Fast Operations for General Loss Functions with Missing Labels}
\label{sec:missing_label}
We assume that the loss function is a general twice-differentiable function $\ell(a,b),$ where 
$a$ and $b$ are scalars. 
Let $\ell'(a,b) = \frac{\partial}{\partial b} \ell(a,b)$, and  
$\ell''(a,b) = \frac{\partial^2}{\partial b^2} \ell(a,b)$. The gradient 
and the Hessian matrix for $g(\w)$ are:\vspace*{-5pt}
\begin{align}
  \nabla g(\w) &= \sum_{(i,j)\in \Omega} \ell'(Y_{ij}, \w^T\tx_{ij})\tx_{ij} + 
  \lambda \w, \label{eq:gw_grad_orig}\\
  \nabla^2 g(\w) &= \sum_{(i,j)\in \Omega} \ell''(Y_{ij}, \w^T\tx_{ij}) 
  \tx_{ij} \tx_{ij}^T + \lambda I \label{eq:gw_hessian_orig}.\vspace*{-5pt}
\end{align}
A direct computation of $\nabla g(\w)$ and $\nabla^2 g(\w) \s$ using 
\eqref{eq:gw_grad_orig} and \eqref{eq:gw_hessian_orig} requires at least 
$O(|\Omega|\bar{d}k)$ time. Below we give faster procedures to 
perform both operations.
\par
\indent{\bf Gradient Calculation.}
Recall that $\tx_{ij} = \h_j \otimes \x_i = \vect{\x_i \h_j^T}$. Therefore, we have
\begin{align*}
  \sum_{(i,j)\in\Omega} \ell'(Y_{ij}, \w^T \tx_{ij}) \x_i \h_j^T = X^T D H,
\end{align*}
where $D$ is sparse with 
$D_{ij} = \ell'(Y_{ij}, \w^T \tx_{ij}),\ \forall (i,j) \in \Omega$. 
Thus, 
\begin{equation}
  \label{eq:gw_grad_missing_fast}
  \nabla g(\w) = \vect{X^TDH} + \lambda \w.
\end{equation}
Assuming that $\ell'(a,b)$ can be computed in constant time, which holds for most 
loss functions (e.g. squared-$L_2$ loss, logistic loss), the gradient 
computation can be done in $O\left(\left(nnz(X)+|\Omega|+d\right)\times k\right)$ time. 
Algorithm \ref{alg:missing} gives the details of computing $\nabla g(\w)$ using 
\eqref{eq:gw_grad_missing_fast}.\\

{\bf Hessian-vector Multiplication.}
After substituting $\tx_{ij} = \h_j \otimes \x_i$, we have
\begin{align*}
  \nabla^2 g(\w) \s = \sum_{(i,j)\in \Omega} \ell''_{ij}\cdot
  \left((\h_j\h_j^T)\otimes (\x_i\x_i^T)\right) \s 
  + \lambda \s, 
\end{align*}
where $\ell''_{ij} = \ell''(Y_{ij}, \w^T\tx_{ij})$. 
Let $S$ be the $d\times k$ matrix such that $\s = \vect{S}$. Using
the identity $(B^T\otimes A) \vect{X} = \vect{AXB}$, we have
$\left((\h_j\h_j^T)\otimes(\x_i\x_i^T)\right)\s = \vect{\x_i\x_i^T S \h_j\h_j^T}$. 
Thus,
\begin{align*}
  \hspace*{-10pt}\sum_{ij} \ell''_{ij} \x_i\x_i^T S \h_j\h_j^T
  &= \sum_{i=1}^n \x_i (\sum_{j:(i,j)\in\Omega} \ell''_{ij}\cdot(S^T\x_i)^T \h_j\h_j^T )\\
  &= \sum_{i=1}^n \x_i (\sum_{j:(i,j)\in\Omega} U_{ij} \h_j^T) = X^TUH,
\end{align*}
where $U$ is sparse, and $U_{ij}~=\ell''_{ij}\cdot(S^T\x_i)^T\h_j,\ \forall (i,j)\in \Omega$.  
Thus, we have 
\begin{equation}
  \label{eq:gw_Hv_missing_fast}
  \nabla^2 g(\w) \s = \vect{X^TUH} + \lambda \s.
\end{equation}
In Algorithm \ref{alg:missing}, we describe a detailed procedure for computing
the Hessian-vector multiplication in \linebreak$O\left(\left(nnz(X)+|\Omega|+d\right)\times k\right)$ time using \eqref{eq:gw_Hv_missing_fast}. \\

  \begin{table*}
    \caption{Computation of $\ell'(a,b)$ and $\ell''(a,b)$ for different loss functions.
    Note that for the logistic and $L_2$-hinge 
loss in Table \ref{tab:grad_loss}, $Y_{ij}$ is 
assumed to be ${-1,+1}$ instead of $\{0,1\}$. 
    }
    \label{tab:grad_loss}
\begin{center}
    \begin{tabular}{llll}
      & $\ell(a,b)$ & $\frac{\partial}{\partial b}\ell(a,b)$ & $\frac{\partial^2}{\partial b^2}\ell(a,b)$ \\
      \hline
      Squared loss & $\frac{1}{2}(a-b)^2$ & $b-a$ & $1$ \\
      Logistic loss & $\log\left(1+e^{-ab}\right)$ & $\frac{-a}{1+e^{-ab}}$ & $\frac{-a^2e^{-ab}}{\left(1+e^{-ab}\right)^2}$ \\
      $L_2$-hinge loss &  
      $\left(\max(0,1-ab)\right)^2$ & $ -2a\max(0,1-ab)$ &  
      $2\cdot\cI[ ab < 1]$
    \end{tabular}
\end{center}
  \end{table*}
{\bf Loss Functions.} See Table \ref{tab:grad_loss} for expressions of $\ell'(a,b)$ and 
$\ell''(a,b)$  for three common loss 
functions: squared loss, logistic loss, and squared hinge loss\footnote{Note that although $L_2$-hinge loss is not twice-differentiable, the sub-differential of $\frac{\partial}{\partial b} \ell(a,b)$ still can be used 
for TRON to solve \eqref{eq:linear_classificaiton}. 
}. Thus, to solve \eqref{eq:linear_classificaiton}, we can apply 
CG for squared loss and the trust region Newton method (TRON)~\cite{CJL07b} for the other two loss functions. 

\subsection{Fast Operations for Squared Loss with Full Labels}
\label{sec:fully_obs}
For the situation where labels are fully observed, solving \eqref{eq:erm} efficiently
in the large-scale setting remains a challenge. The closed form solution from \eqref{eq:close}
is not ideal for two reasons: firstly since it involves the SVD of both $X$ and 
$U_{X}^TY$, the solution becomes infeasible when rank of $X$ is large. Secondly, since it is an unregularized
solution, it might overfit. Indeed \CPLST has similar scalability and overfitting issues due to
absence of regularization and requirement of pseudo inverse calculations for $X$.
When $Y$ is fully observed, Algorithm~\ref{alg:missing}, which aims to handle missing labels with a 
general loss function, is also not scalable as $|\Omega| = nL$ imposing a $O\left(nLk+nnz(X)k\right)$ cost
per operation which is prohibitive when $n$ and $L$ are large.

Although, for a general loss, an $O(nLk)$ cost seems to be inevitable, for 
the $L_2$ loss, we propose fast procedures such that the cost of each operation only 
depends on $nnz(Y)$ instead of $|\Omega|$. In most real-world multi-label 
problems, $nnz(Y) \ll nL = |\Omega|$. As a  
result, for the squared loss, our technique allows alternating minimization to be performed efficiently even when $|\Omega| = nL$. 

If the squared loss is used, the matrix $D$ in Eq. \eqref{eq:gw_grad_missing_fast} 
is $D = XWH^T - Y$ when $Y$ is fully observed, where $W$ is the $d\times k$ 
matrix such that $\vect{W} = \w$. Then, we have 
\begin{equation}
  \nabla g(\w) = \vect{X^TXWH^TH - X^TYH} + \lambda \w.
  \label{eq:gw_grad_full_fast}
\end{equation}
Similarly, $U$ in Eq. \eqref{eq:gw_Hv_missing_fast} is
$U = XSH^T$ which gives us
\begin{equation}
  \nabla^2 g(\w) \s = \vect{X^TX S H^TH} + \lambda \s.
  \label{eq:gw_Hv_full_fast}
\end{equation}
With a careful choice of the sequence of the matrix multiplications, we show 
detailed procedures in Algorithm \ref{alg:full}, which use only $O(nk+k^2)$ 
extra space and $O\left(\left(nnz(Y)+nnz(X)\right)k+ (n+L)k^2\right)$ time to 
compute both $\nabla g(\w)$ and $\nabla^2 g(\w) \s$ efficiently.\\

{\bf Remark on parallelization.}
As we can see, matrix multiplication acts as a crucial subroutine in 
both Algorithms~\ref{alg:missing} and \ref{alg:full}. Thus, 
with a highly-optimized parallel BLAS library (such as ATLAS or Intel MKL), 
our algorithms can easily enjoy speedup brought by the parallel matrix operations 
provided in the library without any extra efforts. Figure 
\ref{fig:speedup} in Appendix~\ref{app:exp} shows that both algorithms do indeed enjoy
impressive speedups as the number of cores increases.\\

{\bf Remark on kernel extension.}
Given a kernel function $\cK(\cdot,\cdot)$, let $f^{j} \in \cH_{\cK}$ be the minimizer of the 
empirical loss defined in Eq. \eqref{eq:erm_missing}. Then by the Representer Theorem \citep[for example,][]{SHS01}, $f^j$
admits a representation of the form: 
  $f^j(\cdot;\z_j) = \sum_{t=1}^{n} z_{jt} \cK(\cdot, \x_t),$ 
where $\z_{j} \in \cR^n$.  Let the vector function 
$\kk(\x): \cR^d \rightarrow \cR^n$ for $\cK$ be defined as 
$\kk(\x) = [\cdots, \cK(\x, \x_t), \cdots]^T$. 
Then $f(\x; Z)$ can be written as $f(\x; Z) = Z^T \kk(\x)$, where 
$Z$ is an $n\times L$ matrix with $\z_j$ as the $j$-th column. 
Once again, we can impose the same trace norm regularization $r(Z)$ and the low rank 
constraint in Eq. \eqref{eq:erm_W_H}. As a result, $Z= WH^T$ and 
$f^j(\x_i, \z_j) = \kk^T(\x_i) W \h_j$. If $K$ is the kernel Gram matrix for 
the training set $\{\x_i:i=1,\ldots,n\}$ and $K_i$ is its $i\th$ column, then the loss in \eqref{eq:erm_W_H} 
can be replaced by $\l(Y_{ij}, K_i^T W \h_j)$. Thus, the proposed alternating 
minimization can be applied to solve Equations~\eqref{eq:erm} and 
\eqref{eq:erm_missing} with the kernel extension as well.


\newcommand{\trr}[1]{\text{tr}\br{#1}}
\newcommand{\norm}[1]{\left\| {#1} \right\|}
\renewcommand{\ind}{\mathds{1}}
\renewcommand{\O}[1]{{\cal O}\br{{#1}}}
\renewcommand{\Z}{{\cal Z}}

\section{Generalization Error Bounds}
\label{sec:bounds}
In this section we analyze excess risk bounds for our learning model with trace norm regularization. Our analysis demonstrates the superiority of our trace norm regularization-based technique over BR and Frobenius norm regularization. We require a more careful analysis for our setting since standard results do not apply because of the presence of missing labels.


Our multi-label learning model is characterized by a distribution $\D$ on the 
space of data points and labels $\X \times \bc{0,1}^L$ where 
$\X \subseteq \R^d$ and a distribution that decides the pattern of missing 
labels. We receive $n$ training points $(\x_1,\y_1),\ldots,(\x_n,\y_n)$ 
sampled i.i.d from the distribution $\D$, where $\y_i \in \bc{0,1}^L$ are 
the \emph{ground truth} label vectors. 
However we shall only be able to observe the ground truth label vectors $\y_i$ 
at $s$ random locations. More specifically, for each $i$ we only observe 
$\y_i$ at locations $l_i^1,\ldots,l_i^s \in [L]$ where the locations are 
chosen uniformly from the set $[L]$ and the choices are independent of 
$(\x_i,\y_i)$.

Given this training data, 
we learn a predictor $\hat Z$ by performing empirical risk minimization over a constrained set of predictors as follows:
\[
\hat Z = \underset{r(Z) \leq \lambda}{\arg\inf}\ \hat\L(Z) = \frac{1}{n}\sum_{i=1}^n\sum_{j=1}^s\ell(\y_i^{l_i^j},f^{l_i^j}(\x_i;Z)),
\]
where $\hat\L(Z)$ is the \emph{empirical risk} of a predictor $Z$. Note that 
although the method in Equation~\ref{eq:erm_missing} uses a regularized 
formulation that is rank-constrained, we analyze just the regularized version 
without the rank constraints for simplicity. As the class of rank-constrained matrices is smaller 
than the class of trace-norm constrained matrices, we can in fact expect 
better generalization performance than that indicated here, if the ERM problem 
can be solved exactly. 

Our goal would be to show that $\hat Z$ has good generalization properties i.e. $
\L(\hat Z) \leq \underset{r(Z) \leq \lambda}{\inf} \L(Z) + \epsilon, $
where $\L(Z) := \EE{\x,\y,l}{\ell(\y^{l},f^{l}(\x;Z))}$ is the \emph{population risk} of a predictor. 
\begin{thm}
\label{thm:genbound-prelim}
Suppose we learn a predictor using the formulation $\hat Z = \underset{\norm{Z}_{\tr} \leq \lambda}{\arg\inf}\ \hat\L(Z)$ over a set of $n$ training points. Then with probability at least $1 - \delta$, we have
\[
\L(\hat Z) \leq \underset{\norm{Z}_{\tr} \leq \lambda}{\arg\inf}\ \L(Z) + \O{s\lambda\sqrt\frac{1}{n}} + \O{s\sqrt\frac{\log\frac{1}{\delta}}{{n}}},
\]
where we assume (w.l.o.g.) that $\E\bsd{\norm{\x}_2^2} \leq 1$.
\end{thm}
We refer to Appendix~\ref{appsec:full=proof} for the proof. Interestingly, we can show that our analysis, obtained via uniform convergence bounds, is tight and cannot be improved in general. We refer the reader to Appendix~\ref{appsec:lower-bound-trace} for the tightness argument. However, it turns out that Frobenius norm regularization is also able to offer the same excess risk bounds and thus, this result does not reveal any advantage for trace norm regularization. Nevertheless, we can still get improved bounds for a general class of distributions over $(\x, \y)$:
\begin{thm}
\label{thm:genbound-main}
Let the data distribution satisfy the following conditions: 1) The top singular value of the covariance matrix $X=\EE{\x \sim \D}{\x\x^\top}$ is $\|X\|_2=\sigma_1$, 2) $\trr{X}=\Sigma$ and 3) 
the distribution on $\X$ is sub-Gaussian i.e. for some $\eta > 0$, for all $\v \in \R^d$,
$
\E\bsd{\exp\left(x^\top \v\right)} \leq \exp\left(\norm{\v}_2^2\eta^2/2\right)
$, then with probability at least $1 - \delta$, we have
\[
\L(\hat Z) \leq \underset{\norm{Z}_{\tr} \leq \lambda}{\arg\inf}\ \L(Z) + \O{s\lambda\sqrt\frac{d(\eta^2 + \sigma_1)}{nL\Sigma}+s\sqrt\frac{\log\frac{1}{\delta}}{{n}}}.
\]
In particular, if the data points are generated from a unit normal distribution, then we have
\[
\L(\hat Z) \leq \underset{\norm{Z}_{\tr} \leq \lambda}{\arg\inf}\ \L(Z) + \O{s\lambda\sqrt\frac{1}{nL}} + \O{s\sqrt\frac{\log\frac{1}{\delta}}{{n}}}.
\]
\end{thm}
The proof of Theorem~\ref{thm:genbound-main} can be found in Appendix~\ref{appsec:full=proof}. Our proofs do not follow either from existing techniques for learning with matrix predictors (for instance \cite{KST12}) or from results on matrix completion with trace norm regularization \cite{SS-S12} due to the complex interplay of feature vectors and missing labels that we encounter in our learning model. Instead, our results utilize a novel form of Rademacher averages, bounding which requires tools from random matrix theory. We note that our results can even handle non-uniform sampling of labels (see Theorem~\ref{thm:genbound-general} in Appendix~\ref{appsec:full=proof} for details).

We note that the assumptions on the data distribution are trivially satisfied with finite $\sigma_1$ and $\eta$ by any distribution with support over a compact set. However, for certain distributions, this allows us to give superior bounds for trace norm regularization.
%
We note that Frobenius norm regularization can give no better than a $\br{\frac{\lambda}{\sqrt n}}$ style excess error bound even for such distributions (see Appendix~\ref{appsec:lower-bound-frob} for a proof), whereas trace norm regularization allows us to get superior $\br{\frac{\lambda}{\sqrt{nL}}}$ style bounds. This is especially contrasting when, for instance, $\lambda = {\cal O}(\sqrt L)$, in which case trace norm regularization gives $\O{\frac{1}{\sqrt n}}$ excess error whereas the excess error for Frobenius regularization deteriorates to $\O{\sqrt\frac{L}{n}}$. Thus, trace norm seems better suited to exploit situations where the data distribution is isotropic. 

Intuitively, we expect such results due to the following reason: when labels are very sparsely observed, such as when $s = \O{1}$, we observe the value of each label on $\O{n/L}$ training points. In such a situation, Frobenius norm regularization with say $\lambda = \sqrt L$ essentially allows an independent predictor $z_l \in \R^d$ to be learned for each label $l \in [L]$. Since all these predictors are being trained on only $\O{n/L}$ training points, the performance accordingly suffers.

On the other hand, if we were to train a single predictor for all the labels i.e. $Z = z\bm{1}^\top$ for some $z \in \R^d$, such a predictor would be able to observe $O(n)$ points and consequently have much better generalization properties. Note that this predictor also satisfies $\|z\bm{1}^\top\|_{\tr} \leq \sqrt L$. This seems to indicate  that trace norm regularization can capture cross label dependencies, especially in the presence of missing labels, much better than Frobenius norm regularization.

Having said that, it is important to note that the two regularizations (trace norm vs. Frobenius norm) might induce different biases in the learning framework. It would be an interesting exercise to study the bias-variance trade-offs offered by these two regularization techniques. However, if one has label correlations then we expect both formulations to suffer similar biases.


\section{Experimental Results}
\label{sec:exp}

\begin{table}
  \caption{Data statistics. $d$ and $L$ are the number of features and labels, 
  respectively, and $\bar{d}$ and $\bar{L}$ are the average number of nonzero 
  features and positive labels in an instance, respectively.}  
  \label{tab:data-stat}
  \centering
  {
  \begin{tabular}{l|rr|rrr|rrr}
    & & & \multicolumn{3}{c}{Training set} & \multicolumn{3}{|c}{Test set} \\
    Dataset & $d$ &$L$& $n$ & $\bar{d}$&  
    $\bar{L}$&
    $n$ & $\bar{d}$& $\bar{L}$\\ 
    \hline
    bibtex & 1,836 & 159& 4,880 & 68.74 & 2.40& 2,515 & 68.50 & 2.40 \\
    autofood & 9,382 & 162& 155 & 143.92 & 15.80& 38 & 143.71 & 13.71 \\
    compphys & 33,284 & 208& 161 & 792.78 & 9.80& 40 & 899.02  & 11.83 \\
    delicious & 500 & 983& 12,920 & 18.17 & 19.03 & 3,185 & 18.80 & 19.00 \\
    eurlex & 5,000 & 3,993& 17,413 & 236.69 & 5.30& 1,935 & 240.96 & 5.32 \\
    nus-wide & 1,134& 1,000& 161,789& 862.70 & 5.78& 107,859 & 862.94  & 5.79 \\
    wiki & 366,932& 213,707& 881,805& 146.78 & 7.06& 10,000 & 147.78  & 7.08 \\
  \end{tabular}
  }
\end{table}

We now present experimental results in order to assess our 
proposed algorithms in terms of accuracy and scalability. As we shall
see, the results unambiguously demonstrate the superiority of our method
over other approaches.\\


{\bf Datasets.}
We consider a variety of benchmark datasets including four standard 
datasets (bibtex, delicious, eurlex, and nus-wide), two datasets with $d \gg L$  
(autofood and compphys), and a very large scale Wikipedia based dataset,
which contains about 1M wikipages and 200K labels.
See Table \ref{tab:data-stat} for more information about the datasets. We conducted all experiments on
an Intel machine with 32 cores. \\

\begin{table}[h]
  \caption{Comparison of \MLR with various loss functions and \WSABIE on 
  smaller datasets. SQ denotes squared loss, LR denotes logistic regression loss, and 
  SH denotes squared hinge loss}  
  \label{tab:full_label_var_loss}
  \begin{center}
    {

    \begin{tabular}{@{}l@{}r|rrr|r||rrr|r@{}}
      &	& \multicolumn{4}{@{}c@{}}{Top-$3$ Accuracy} & 
      \multicolumn{4}{@{}c@{}}{Average AUC}\\
      &	& \multicolumn{3}{@{}c@{}}{\MLR} &\multirow{2}{*}{{\scriptsize \WSABIE}} & 
      \multicolumn{3}{@{}|c@{}}{\MLR} & \multirow{2}{*}{{\scriptsize \WSABIE}} \\
      &$k/L$ & SQ & LR & SH & & SQ & LR & SH &\\
      \hline
      \hline
      \multirow{3}{*}{bibtex} 
      &	$20\%$ & {\bf 34.16}& 25.65& 27.37& 28.77& 0.8910& 0.8677& 0.8541& {\bf 0.9055}\\
      &	$40\%$ & {\bf 36.53}& 28.20& 24.81& 30.05& 0.9015& 0.8809& 0.8467& {\bf 0.9092}\\
      &	$60\%$ & {\bf 38.00}& 28.68& 23.26& 31.11& 0.9040& 0.8861& 0.8505& {\bf 0.9089}\\
      \hline
      \multirow{3}{*}{autofood} 
      &	$20\%$ & {\bf 81.58}& 80.70& {\bf 81.58}& 66.67& 0.9565& {\bf 0.9598}& 0.9424& 0.8779\\
      &	$40\%$ & 76.32& {\bf 80.70}& 78.95& 70.18&       0.9277& {\bf 0.9590}& 0.9485& 0.8806\\
      &	$60\%$ & 70.18& 80.70& {\bf 81.58}& 60.53&       0.8815& {\bf 0.9582}& 0.9513& 0.8518\\
      \hline
      \multirow{3}{*}{compphys} 
      &	$20\%$ & {\bf 80.00}& {\bf 80.00}& {\bf 80.00}& 49.17& 0.9163& 0.9223& {\bf 0.9274}& 0.8212\\
      &	$40\%$ & {\bf 80.00}& 78.33& 79.17& 39.17&             {\bf 0.9199}& 0.9157& 0.9191& 0.8066\\
      &	$60\%$ & {\bf 80.00}& {\bf 80.00}& {\bf 80.00}& 49.17& {\bf 0.9179}& 0.9143& 0.9098& 0.8040\\
      \hline
      \multirow{3}{*}{delicious} 
      &	$20\%$ & {\bf 61.20}& 53.68& 57.27& 42.87& 0.8854& 0.8588& {\bf 0.8894}& 0.8561\\
      &	$40\%$ & {\bf 61.23}& 49.13& 52.95& 42.05& 0.8827& 0.8534& {\bf 0.8868}& 0.8553\\
      &	$60\%$ & {\bf 61.15}& 46.76& 49.58& 42.22& 0.8814& 0.8517& {\bf 0.8852}& 0.8523\\
    \end{tabular}
    }
  \end{center}
\end{table}

\begin{table}[h]
    \caption{Comparison of \MLR and \WSABIE on large datasets}
  \label{tab:full-label-large-set}
  \begin{center}
    {
    \begin{tabular}{@{}lrrrrr|rrrr@{}}
      &  & \multicolumn{4}{c|}{\MLR} & \multicolumn{4}{c}{\WSABIE} \\
      dataset   & $k$ & time (s) &top-$1$ & top-$3$ & AUC & time (s) & top-$1$ & top-$3$ & AUC \\
      \hline
      \hline
      \multirow{2}{*}{eurlex} 
      & 250 & {\bf 175} & {\bf 51.99} & {\bf 39.79} & {\bf 0.9425} & 373& 33.13 & 25.01 &  0.8648 \\
      & 500 & {\bf 487}& {\bf 56.90} & {\bf 44.20} & {\bf 0.9456} & 777 & 31.58 & 24.00 & 0.8651 \\
      \cline{1-10}
      \multirow{2}{*}{nus-wide} 
      & 50 & {\bf 574} & {\bf 20.71} & {\bf 15.96} & {\bf 0.7741} & 4,705 & 14.58 & 11.37 & 0.7658 \\
      & 100 & {\bf 1,097} & {\bf 20.76} & {\bf 16.00} & {\bf 0.7718} & 6,880 & 12.46 & 10.21 & 0.7597 \\
      \cline{1-10}
      \multirow{2}{*}{wiki} 
      & 250 & {\bf 9,932} & {\bf 19.56} & 14.43 & {\bf 0.9086} & 79,086 & 18.91 & {\bf 14.65} & 0.9020 \\
      & 500 & {\bf 18,072}& {\bf 22.83} & {\bf 17.30} & {\bf 0.9374} & 139,290 & 19.20 & 15.66 & 0.9058 \\
    \end{tabular}
    }
\end{center}
\end{table}

{\bf Competing Methods.}
A list containing details of the competing methods (including ours) is given below.
Note that \CS~\cite{DH09a} and 
\PLST~\cite{FT12a} are not included as they are shown to be suboptimal to \CPLST and 
\BCS in \cite{YNC12a,AK12a}.  
\begin{compactitem}[]
  \item[1.] \MLR ({\bf L}ow rank {\bf E}mpirical risk minimization for 
    {\bf M}ulti-Label {\bf L}earning): our proposed method.  We implemented CG with Algorithms 
    \ref{alg:missing} and \ref{alg:full} for squared loss, and 
    TRON~\cite{CJL07b} with Algorithm \ref{alg:missing} for logistic and 
    squared hinge loss.  
  \item[2.] \CPLST: the method proposed in \cite{YNC12a}. We used the code provided by the author.
  \item[3.] \BCS: the Bayesian compressed sensing method of \cite{AK12a}. We used code provided by the authors to test this method.
  \item[4.] \BR: Binary Relevance with various loss functions.  
  \item[5.] \WSABIE: As there is no publicly available code, we implemented this
    method and hand-tuned the learning rate and the margin for each dataset as 
    suggested by the authors of \WSABIE~\citep{mailJW13}.\\
\end{compactitem}

{\bf Evaluation Criteria.} 
Three criteria will be used to compare the methods: top-K 
accuracy, which measures the performance on a few top predictions, Hamming 
loss, which considers the overall classification performance, and average AUC, 
which takes the overall ranking into account. 

Given a test set $\{\x_i, \y_i: i = 1,\ldots,n\}$ and an real-valued predictor 
$f(\x):\R^d\rightarrow \R$:   
\begin{itemize}
  \item Top-$K$ accuracy: for each instance, we select the $K$ labels with 
    the largest decision values for prediction. The average accuracy among all 
    instances are reported as the top-$K$ accuracy.
  \item Hamming-loss: for each pair of instance $\x$ and label index $j$, we round the 
    decision value $f^j(\x)$ to 0 or 1. 
    \[
    \text{Hamming Loss} = \frac{1}{nL}\sum_{i=1}^n \sum_{j=1}^L I[\text{round}\left(f^j(\x)\right) \neq \y^j]
    \]
  \item Average AUC: we follow \cite{SSB09a} to calculate area under ROC curve 
    for each instance and report the average AUC among all test instances.  
\end{itemize}
\subsection{Results with full labels}
We divide datasets into two groups: \emph{small datasets} (bibtex, autofood, compphys, and
delicious) on which \BCS and \CPLST can be tested without scalability issues 
and \emph{large datasets} (eurlex, nus-wide, and wiki) to which only \MLR and 
\WSABIE are able to scale.\\

\begin{figure}
	\centering
  \begin{tabular}{@{}c@{}c@{}c@{}c}
    \hspace{-1.3em}
    \subfloat[bibtex]{
    \label{fig:full_label_bibtex}
    \includegraphics[width=0.37\linewidth]{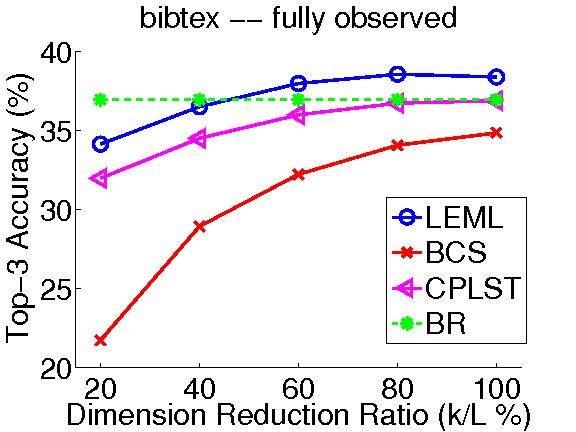}
    } &
    \hspace{-1.5em}
   \subfloat[compphys]{
   \label{fig:full_label_compphys}
   \includegraphics[width=0.37\linewidth]{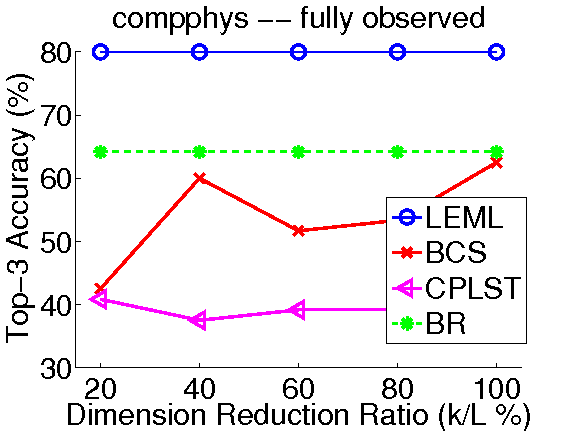}
   }  & 
    \hspace{-1.5em}
   \subfloat[delicious]{
   \label{fig:full_label_delicious}
   \includegraphics[width=0.37\linewidth]{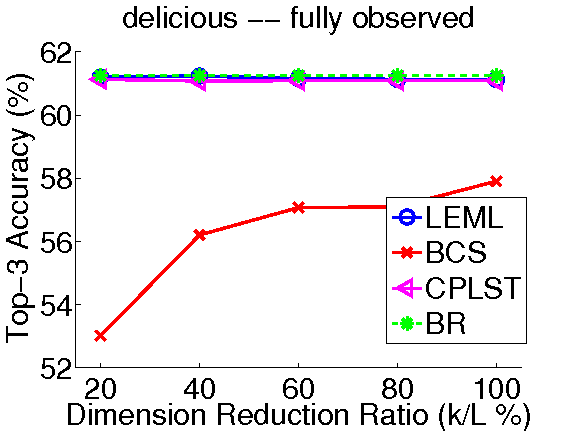}
   } 
  \end{tabular}
  \caption{Comparison between different dimensionality reduction methods with 
  fully observed $Y$ by varying the ratio of dimension reduction $k/L$.}
  \label{fig:full_label}
\end{figure}

{\bf Small datasets.} We first compare dimension reduction based approaches
to assess their performance with varying dimensionality reduction ratios.
Figure~\ref{fig:full_label} presents these results for \MLR, \CPLST and \BCS
on the squared $L_2$ loss with \BR included for reference. Clearly \MLR consistently
performs better than other methods for all ratios. Next we compare \MLR with three loss functions 
(squared, logistic, and $L_2$-hinge) to \WSABIE, which approximately optimizes 
a weighted approximate ranking loss. As Table \ref{tab:full_label_var_loss} shows, although the best 
loss for each dataset might vary, \MLR is always superior to or competitive with \WSABIE.
Based on Figure \ref{fig:full_label}, Table \ref{tab:full_label_var_loss}, 
and further results in Appendix \ref{app:exp-full}, we make the following 
observations. 1) \MLR can deliver accuracies competitive with \BR even with a
severe reduction in dimensionality,  
2) On bibtex and compphys, \MLR is even shown to outperform \BR. This is a 
benefit brought forward by the design of \MLR, wherein the relation between labels 
can be captured by a low rank $Z$. This enables \MLR to better utilize 
label information than \BR and yield better accuracies. 
3) On autofood and compphys, \CPLST 
seems to suffer from overfitting and demonstrates a significant dip in performance.
In contrast, \MLR, which brings regularization into the formulation performs well 
consistently on all datasets. \\

{\bf Larger data.} Table \ref{tab:full-label-large-set} shows results for 
\MLR and \WSABIE on the three larger datasets. We implemented \MLR with the 
squared $L_2$ loss using Algorithm \ref{alg:full} for comparison in the full labels case.
Note that Hamming loss is not used here as it is not 
clear how to convert the label ranking given by \WSABIE to a 0/1 encoding. 
For \MLR, we report the time and the accuracies obtained after five alternating 
iterations. For \WSABIE, we ran the method on each dataset with 
the hand-tuned parameters for about two days, and reported the time and results 
for the epoch with the highest average AUC.  
On eurlex and nus-wide, \MLR is clearly superior than \WSABIE on all 
evaluation criteria. On wiki, although both methods share a similar performance
for $k=250$, on increasing $k$ to $500$, \MLR again outperforms \WSABIE. Also clearly
noticeable is the stark difference in the running times of the two methods.
Whereas \MLR takes less than 6 hours to deliver 0.9374 AUC on wiki, \WSABIE requires 
about 1.6 days to achieve 0.9058 AUC. More specifically, \WSABIE takes 
about 7,000s for the first epoch, 16,000s for the second and 36,000s for the third epoch 
which result in it spending almost two days on just 5 epochs. Although this phenomenon 
is expected due to the sampling scheme in \WSABIE~\cite{JW10a}, it becomes 
more serious as $L$ increases. We leave the issue of designing a better sampling 
scheme with large $L$ for future work. Figure~\ref{fig:time-vs-auc} further illustrates
this gap in training times for the nus-wide dataset. All in all, the results clearly demonstrate the 
scalability and efficiency of \MLR.


\begin{table}
\caption{Comparison between various dimensionality reduction approaches on $Y$ 
with 20\% observed entries, and $k=0.4 L$.}
\label{tab:missing_square_loss}
\centering
{
\begin{tabular}{@{}l@{}|c@{ }c@{ }c@{}||c@{ }c@{ }c@{}||c@{ }c@{ }c@{}}
  & \multicolumn{3}{@{}c@{}}{Top-$3$ Accuracy} 
  &\multicolumn{3}{@{}c@{}}{Hamming loss}   
  & \multicolumn{3}{@{}c@{}}{Average AUC}  \\
  & \MLR& \BCS& \BR& \MLR& \BCS& \BR& \MLR& \BCS& \BR\\
  \hline
  bibtex& {\bf 28.50}& 23.84& 25.78& {\bf 0.0136}&  0.2496& 0.0193 & {\bf 0.8332} & 0.7871 & 0.8087\\
  autofood& {\bf 67.54}& 35.09&  62.28& {\bf 0.0671}& 0.2445& 0.0760& {\bf 0.8634} & 0.6322 & 0.8178 \\
  compphys& {\bf 65.00}& 35.83& 31.67& {\bf 0.0518}&  0.2569& 0.0566& {\bf 0.7964} & 0.6442 & 0.7459 \\
\end{tabular}
}
\end{table}
\subsection{Results with missing labels}
\begin{figure}
	\centering
  \begin{tabular}{@{}c@{}c@{}c@{}c}
    \hspace{-1.3em}
     \subfloat[]{
     \label{fig:time-vs-auc}
     \includegraphics[width=0.37\linewidth]{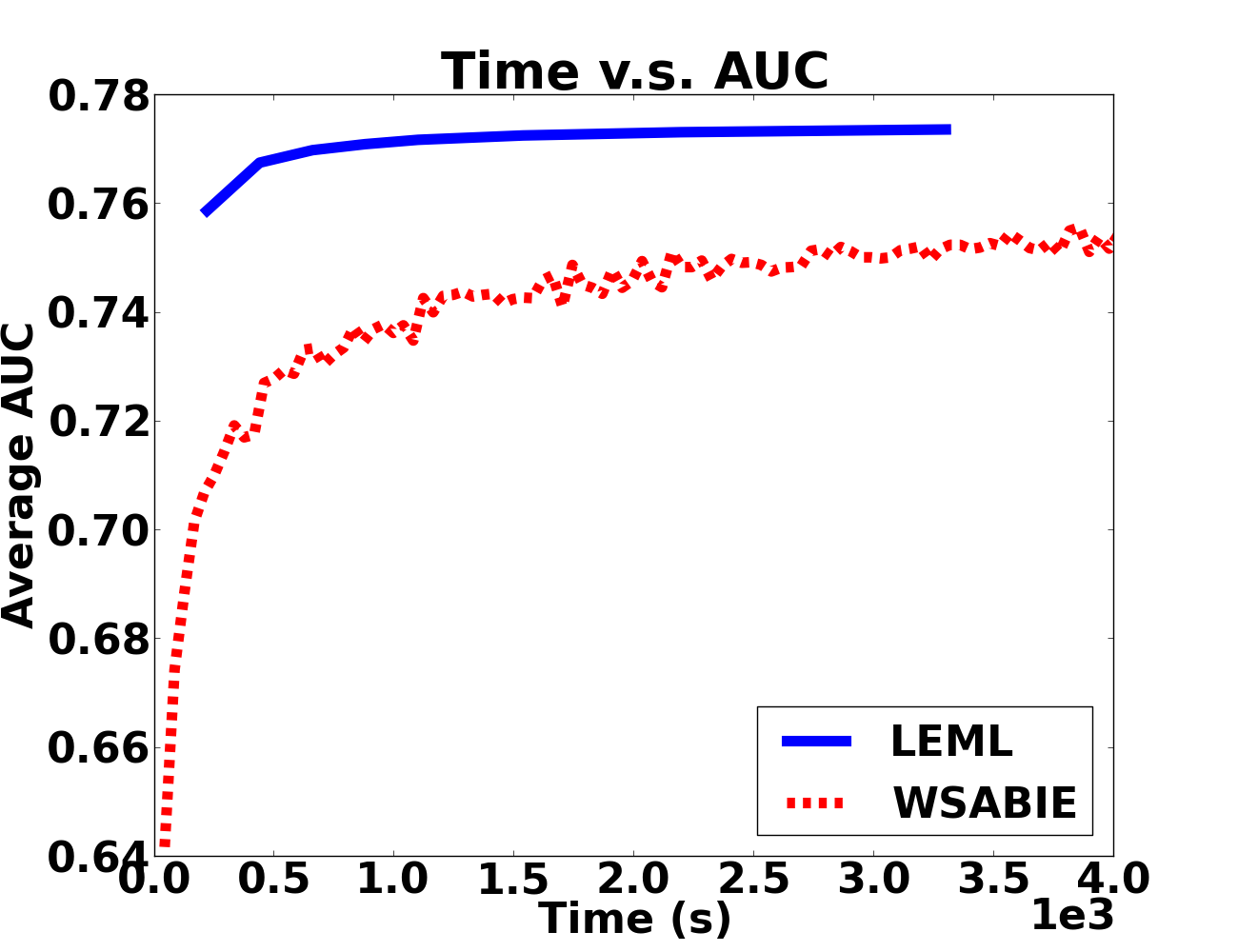}
     }  &
    \hspace{-1.5em}
    \subfloat[]{
    \label{fig:fix_rank}
    \includegraphics[width=0.37\linewidth]{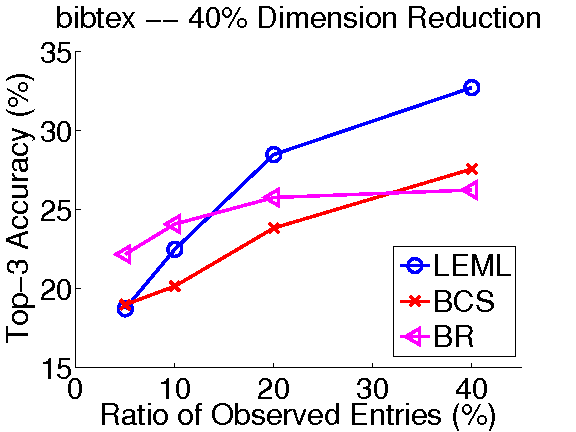}
    } &
    \hspace{-1.42em}
   \subfloat[]{
   \label{fig:fix_obs}
   \includegraphics[width=0.37\linewidth]{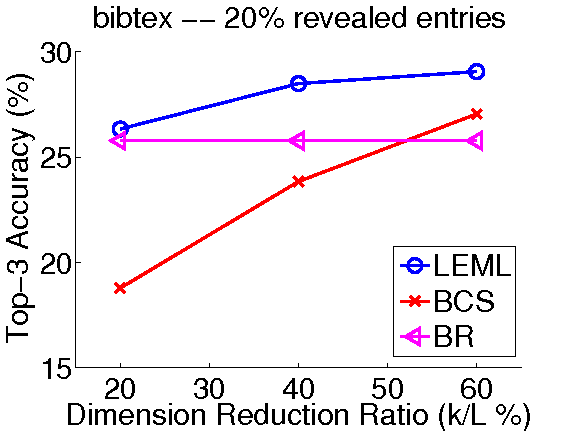}
   }   
  \end{tabular}
  \caption{Results for (a): \MLR vs. \WSABIE on nus-wide. (b): bibtex with various observed 
  ratios. (c): bibtex with various ratios of dimension reduction.}
\end{figure}

For experiments with missing labels, we 
compare \MLR, \BCS, and \BR. We implemented \BR with missing labels by learning 
an $L_2$-regularized binary classifier/regressor for each label on observed 
instances. Thus, the model derived from \BR corresponds to the minimizer of 
\eqref{eq:erm_missing} with Frobenius norm regularization. 
Table \ref{tab:missing_square_loss} shows the results when $20\%$ entries were revealed
(i.e. $80\%$ missing rate) and squared loss function was used for training.
We used $k=0.4L$ for both \MLR and \BCS. The results clearly show that \MLR outperforms 
\BCS and \MLR with respect to all three evaluation criteria. On bibtex, we 
further present results for various rates of observed labels in Figure 
\ref{fig:fix_rank} and results for various dimension reduction ratios in 
Figure \ref{fig:fix_obs}. \MLR clearly shows superior performance over other approaches, 
which corroborates the theoretical results of Section~\ref{sec:bounds} that 
indicate better generalization performance for low-rank promoting regularizations. More empirical 
results for other loss functions, various observed ratios and dimension reduction ratios can be found in Appendix \ref{app:exp-missing}.

\section{Conclusion}
\label{sec:conclusion}
In this paper we studied the multi-label learning problem with missing labels in the standard ERM framework. We modeled our framework with rank constraints and regularizers to increase scalability and efficiency. To solve the obtained non-convex problem, we proposed an alternating minimization based method that critically exploits structure in the loss function to make our method scalable. 
We showed that our learning framework admits excess risk bounds that indicate  better generalization performance for our methods than the existing methods like BR, something which our experiments also confirmed. Our experiments additionally demonstrated that our techniques are much more efficient than other large scale multi-label classifiers and give superior performance than the existing label compression based approaches. For future work, we would like to extend \MLR to other (non decomposable) loss functions such as ranking losses and study conditions under which alternating minimization for our problem is guaranteed to converge to the global optimum. Another open question is if our risk bounds can be improved by avoiding the uniform convergence route that we use in the paper.


\bibliographystyle{icml2014}
\bibliography{refs}
\clearpage
\newpage
\appendix
\onecolumn
\allowdisplaybreaks
\section{Algorithm Details}
\label{app:detailed_alg}
  \subsection{Conjugate Gradient for Squared Loss }
\label{app:detailed_alg}
In Algorithm \ref{alg:cg}, we show the detailed conjugate gradient procedure 
used to solve \eqref{eq:linear_classificaiton} when the squared loss is used. 
Note that $\nabla^2 g(\w)$ is invariant to $\w$ as 
\eqref{eq:linear_classificaiton} is a quadratic problem due to the squared 
loss function. 

\begin{Algorithm}[h]
  \caption{Conjugate gradient for solving \eqref{eq:linear_classificaiton} 
  with the squared loss}
  \label{alg:cg}
  \begin{compactitem}
  \item Set initial $\w_0$, $\r_0 = - \nabla g(\w_0)$, $\d_0 = \r_0$.
  \item For $t=0,1,2,\ldots$
    \begin{compactitem}
    \item If $\|\r_t\|$ is small enough, then stop the procedure and return $\w_t$.
    \item $\displaystyle\alpha_t = \frac{\r_t^T\r_t}{\d_t^T \nabla^2 g(\w_0) \d_t}$
    \item $\w_{t+1} = \w_t + \alpha_t \d_t$
    \item $\r_{t+1} = \r_{t} - \alpha_t \nabla^2 g(\w_0) \d_t$
    \item $\displaystyle\beta_{t} = \frac{\r_{t+1}^T \r_{t+1}}{\r_t^T \r_t}$
    \item $\d_{t+1} = \r_{t+1} + \beta_{t} \d_t$
    \end{compactitem}
  \end{compactitem}
\end{Algorithm}

\newcommand{\Om}[1]{\Omega\br{{#1}}}
\renewcommand{\E}[1]{{\mathbb E}\bsd{{#1}}}
\newcommand{\vv}{\vec{v}}

\section{Analyzing Trace Norm-bounded Predictors}
\label{appsec:full=proof}

In this section, we shall provide a proof of Theorems~\ref{thm:genbound-prelim} and \ref{thm:genbound-main}. Our proof shall proceed by demonstrating a uniform convergence style bound for the empirical losses. More precisely, we shall show, for both trace norm as well as Frobenius regularizations, that with high probability, we have
\[
\L(\hat Z) \leq \hat\L(\hat Z) + \epsilon.
\]
Suppose $Z^\ast \in \underset{r(Z) \leq \lambda}{\arg\min}\ \L(Z)$, then a similar analysis will allow us to show, again with high probability,
\[
\hat\L(Z^\ast) \leq \L(Z^\ast) + \epsilon.
\]
Combining the two along with the fact that $\hat Z$ is the empirical risk minimizer i.e. $\hat\L(\hat Z) \leq \hat\L(Z^\ast)$ will yield the announced claim in the following form:
\[
\L(\hat Z) \leq \L(Z^\ast) + 2\epsilon.
\]

Thus, in the sequel, we shall only concentrate on proving the aforementioned uniform convergence bound. We shall denote the regularized class of predictors as $\Z = \bc{Z \in \R^{d \times L}, r(Z) \leq \lambda}$, where $r(Z) = \norm{Z}_{\tr}$ or $r(Z) = \norm{Z}_F$. We shall also use the following shorthand for the loss incurred by the predictor on a specific label $l \in [L]$: $\ell(\y_i^{l},Z_l,\x) := \ell(\y_i^{l},f^{l}(\x;Z))$, where $Z_l$ denotes the $l\th$ column of the matrix $Z$.

We shall perform our analysis in several steps outlined below:
\begin{enumerate}
	\item Step 1: In this step we shall show, by an application of McDiarmid's inequality, that with high probability, the excess risk of the learned predictor can be bounded by bounding the expected supr\=emus deviation of empirical risks from population risks over the set of predictors in the class $\Z$.
	\item Step 2: In this step we shall show that the expected supr\=emus deviation can be bounded by a Rademacher average term.
	\item Step 3: In this step we shall reduce the estimation of the Rademacher average term to the estimation of the spectral norm of a random matrix that we shall describe.
	\item Step 4: Finally, we shall use tools from random matrix theory to bound the spectral norm of the random matrix.
\end{enumerate}
We now give details of each of the steps in the following subsections:

\subsection{Step 1: Bounding Excess Risk by Expected Supr\=emus Deviation}
We will first analyze the case $s=1$ and will later show how to extend the analysis to $s > 1$. In this case, we receive $n$ training points $(\x_i,\y_i)$ and for each training point $\x_i$, we get to see the value of a random label $l_i \in [L]$ i.e. we get to see the true value of $\y_i^{l_i}$. Thus, for any predictor $Z \in \Z$, the observed training loss is given by
\[
\hat\L(Z) = \frac{1}{n}\sum_{i=1}^n\ell(\y_i^{l_i},Z_{l_i},\x_i).
\]
The population risk functional is given by
\[
\L(Z) = \EE{(\x,\y,l)}{\ell_l(\y^l,f^l(\x;Z))} = \EE{(\x,\y,l)}{\ell_l(\y^l,Z_l,\x)}
\]
We note here that our subsequent analysis shall hold even for non uniform distributions for sampling the labels. The definition of the population risk functional incorporates this. In case we have a uniform distribution over the labels, the above definition reduces to
\[
\L(Z) = \EE{(\x,\y,l)}{\ell_l(\y^l,Z_l,\x)} = \EE{(\tilde \x_i,\tilde \y_i,\tilde l_i)}{\frac{1}{n}\sum_{i=1}^n\ell(\tilde \y_i^{\tilde l_i},Z_{\tilde l_i},\tilde \x_i)}
\]

Given the above, we now analyze the excess risk i.e. the difference between the observed training loss $\hat\L(\hat Z)$ and the population risk $\L(\hat Z)$. 
\begin{eqnarray*}
\L(\hat Z) - \hat\L(\hat Z) &\leq& \usup{Z \in \Z}\bc{\L(Z) - \hat\L(Z)}\\
							&=& \underbrace{\usup{Z \in \Z}\bc{\EE{(\tilde \x_i,\tilde \y_i,\tilde l_i)}{\frac{1}{n}\sum_{i=1}^n\ell(\tilde \y_i^{\tilde l_i},Z_{\tilde l_i},\tilde \x_i)} - \frac{1}{n}\sum_{i=1}^n\ell(\y_i^{l_i},Z_{l_i},\x_i)}}_{g((\x_1,\y_1,l_1),\ldots,(\x_n,\y_n,l_n))}
\end{eqnarray*}
Since all the label-wise loss functions are bounded, an arbitrary change in any $(\x_i,\y_i)$ or any $l_i$ should not perturb the expression $g((\x_1,\y_1,l_1),\ldots,(\x_n,\y_n,l_n))$ by more than $\O{\frac{1}{n}}$. Thus, by an application of McDiarmid's inequality, we have, with probability at least $1 - \delta$,
\[
\L(\hat Z) - \hat\L(\hat Z) \leq \EE{(\x_i,\y_i),l_i}{g((\x_1,\y_1,l_1),\ldots,(\x_n,\y_n,l_n))} + \O{\sqrt\frac{\log\frac{1}{\delta}}{n}}
\]
Thus, we conclude that the excess risk of the learned predictor can be bounded by calculating the expected supr\=emus deviation of empirical risks from population risks.

\subsection{Step 2: Bounding Expected Supr\=emus Deviation by a Rademacher Average}
We now analyze the expected supr\=emus deviation. We have
\begin{eqnarray*}
&& \EE{(\x_i,\y_i),l_i}{g((\x_1,\y_1,l_1),\ldots,(\x_n,\y_n,l_n))}\\
&=& \EE{(\x_i,\y_i),l_i}{\usup{Z \in \Z}\bc{\EE{(\tilde \x_i,\tilde \y_i,\tilde l_i)}{\frac{1}{n}\sum_{i=1}^n\ell(\tilde \y_i^{\tilde l_i},Z_{\tilde l_i},\tilde \x_i)} - \frac{1}{n}\sum_{i=1}^n\ell(\y_i^{l_i},Z_{l_i},\x_i)}}\\
&\leq& \EE{(\x_i,\y_i),l_i}{\usup{Z \in \Z}\bc{\EE{(\tilde \x_i,\tilde \y_i,\tilde l_i)}{\frac{1}{n}\sum_{i=1}^n\ell(\tilde \y_i^{\tilde l_i},Z_{\tilde l_i},\tilde \x_i)} - \frac{1}{n}\sum_{i=1}^n\EE{(\tilde \x_i,\tilde \y_i)}{\ell(\tilde \y_i^{l_i},Z_{l_i},\tilde \x_i)}}}\\
&&+\ \EE{(\x_i,\y_i),l_i}{\usup{Z \in \Z}\bc{\frac{1}{n}\sum_{i=1}^n\EE{(\tilde \x_i,\tilde \y_i)}{\ell(\tilde \y_i^{l_i},Z_{l_i},\tilde \x_i)} - \frac{1}{n}\sum_{i=1}^n\ell(\y_i^{l_i},Z_{l_i},\x_i)}}\\
&=& \EE{l_i}{\usup{Z \in \Z}\bc{\EE{(\tilde \x_i,\tilde \y_i,\tilde l_i)}{\frac{1}{n}\sum_{i=1}^n\ell(\tilde \y_i^{\tilde l_i},Z_{\tilde l_i},\tilde \x_i)} - \frac{1}{n}\sum_{i=1}^n\EE{(\tilde \x_i,\tilde \y_i)}{\ell(\tilde \y_i^{l_i},Z_{l_i},\tilde \x_i)}}}\\
&&+\ \EE{(\x_i,\y_i),l_i}{\usup{Z \in \Z}\bc{\frac{1}{n}\sum_{i=1}^n\EE{(\tilde \x_i,\tilde \y_i)}{\ell(\tilde \y_i^{l_i},Z_{l_i},\tilde \x_i)} - \frac{1}{n}\sum_{i=1}^n\ell(\y_i^{l_i},Z_{l_i},\x_i)}}\\
&\leq& \EE{(l_i,\tilde l_i)}{\usup{Z \in \Z}\bc{\frac{1}{n}\sum_{i=1}^n\EE{(\tilde \x_i,\tilde \y_i)}{\ell(\tilde \y_i^{\tilde l_i},Z_{\tilde l_i},\tilde \x_i)} - \frac{1}{n}\sum_{i=1}^n\EE{(\tilde \x_i,\tilde \y_i)}{\ell(\tilde \y_i^{l_i},Z_{l_i},\tilde \x_i)}}}\\
&&+\ \EE{(\x_i,\y_i),l_i,(\tilde \x_i, \tilde \y_i)}{\usup{Z \in \Z}\bc{\frac{1}{n}\sum_{i=1}^n{\ell(\tilde \y_i^{l_i},Z_{l_i},\tilde \x_i)} - \frac{1}{n}\sum_{i=1}^n\ell(\y_i^{l_i},Z_{l_i},\x_i)}}\\
&=& \EE{(l_i,\tilde l_i),\epsilon_i}{\usup{Z \in \Z}\bc{\frac{1}{n}\sum_{i=1}^n\epsilon_i\br{\EE{(\tilde \x_i,\tilde \y_i)}{\ell(\tilde \y_i^{\tilde l_i},Z_{\tilde l_i},\tilde \x_i)} - \EE{(\tilde \x_i,\tilde \y_i)}{\ell(\tilde \y_i^{l_i},Z_{l_i},\tilde \x_i)}}}}\\
&&+\ \EE{(\x_i,\y_i),l_i,(\tilde \x_i, \tilde \y_i),\epsilon_i}{\usup{Z \in \Z}\bc{\frac{1}{n}\sum_{i=1}^n\epsilon_i\br{\ell(\tilde \y_i^{l_i},Z_{l_i},\tilde \x_i) - \ell(\y_i^{l_i},Z_{l_i},\x_i)}}}\\
&\leq& 2\EE{l_i,\epsilon_i}{\usup{Z \in \Z}\bc{\frac{1}{n}\sum_{i=1}^n\epsilon_i{\EE{(\tilde \x_i,\tilde \y_i)}{\ell(\tilde \y_i^{l_i},Z_{l_i},\tilde \x_i)}}}} + 2\EE{(\x_i,\y_i),l_i,\epsilon_i}{\usup{Z \in \Z}\bc{\frac{1}{n}\sum_{i=1}^n\epsilon_i{\ell(\y_i^{l_i},Z_{l_i},\x_i)}}}\\
&\leq& 2\EE{(\tilde \x_i,\tilde \y_i),l_i,\epsilon_i}{\usup{Z \in \Z}\bc{\frac{1}{n}\sum_{i=1}^n\epsilon_i{{\ell(\tilde \y_i^{l_i},Z_{l_i},\tilde \x_i)}}}} + 2\EE{(\x_i,\y_i),l_i,\epsilon_i}{\usup{Z \in \Z}\bc{\frac{1}{n}\sum_{i=1}^n\epsilon_i{\ell(\y_i^{l_i},Z_{l_i},\x_i)}}}\\
&\leq& 4\EE{(\x_i,\y_i),l_i,\epsilon_i}{\usup{Z \in \Z}\bc{\frac{1}{n}\sum_{i=1}^n\epsilon_i{{\ell(\y_i^{l_i},Z_{l_i},\x_i)}}}} \leq \frac{4C}{n}\EE{(\x_i,\y_i),l_i,\epsilon_i}{\usup{Z \in \Z}\bc{\sum_{i=1}^n\epsilon_i{{\ip{Z_{l_i}}{\x_i}}}}}\\
&=& \frac{4C}{n}\EE{X,\vecl,\veceps}{\usup{Z\in\Z}{\ip{Z}{X^{\vecl}_\veceps}}},
\end{eqnarray*}
where for any $\x_1,\ldots,\x_n \in \X$, $\vecl \in [L]^{n}$ and $\veceps \in \bc{-1,+1}^{n}$, we define the matrix $X^\vecl_\veceps$ as follows:
\[
X^\vecl_\veceps := \bs{\sum_{i \in I_1}\epsilon_i\x_i \sum_{i \in I_2}\epsilon_i\x_i \ldots \sum_{i \in I_L}\epsilon_i\x_i}
\]
where for any  $l \in [L]$, we define $I_l := \bc{i: l_i = l}$. Note that in the last second inequality we have used the contraction inequality for Rademacher averages \citep[see][proof of Theorem 4.12]{LT02} We also note that the above analysis also allows for separate label-wise loss functions, so long as they are all bounded and $C$-Lipschitz. For any matrix predictor class $\Z$, we define	its Rademacher complexity as follows:
\[
\RR_{n}\br{\Z} := \frac{1}{n}\EE{X,\vecl,\veceps}{\usup{Z\in\Z}{\ip{Z}{X^{\vecl}_\veceps}}}
\]
We have thus established that with high probability,
\[
\L(\hat Z) - \hat\L(\hat Z) \leq 4C\RR_{n}\br{\Z} + \O{\sqrt\frac{\log\frac{1}{\delta}}{n}}.
\]
We now establish that the same analysis also extends to situations wherein, for each training point we observe values of $s$ labels instead. Thus, for each $\x_i$, we observe values for labels $l_i^1,\ldots,l_i^s$. In this case the empirical loss is given by
\[
\hat\L(Z) = \frac{1}{n}\sum_{i=1}^{n}\sum_{j=1}^s\ell(\y_i^{l^j_i},Z_{l^j_i},\x_i)
\]
The change in any $\x_i$ leads to a perturbation of at most $\O{\frac{s}{n}}$ whereas the change in any $l^j_i$ leads to a perturbation of $\O{\frac{1}{n}}$. Thus the sum of squared perturbations is bounded by $\frac{2s^2}{n}$. Thus on application of the McDiarmid's inequality, we will be able to bound the excess risk by the following expected supr\=emus deviation term
\[
\EE{(\x_i,\y_i,l^j_i)}{\usup{Z\in\Z}\bc{s\EE{(x,y,l)}{\ell_l(\y^l,Z_l,\x)} - \frac{1}{n}\sum_{i=1}^{n}\sum_{j=1}^s\ell(\y_i^{l^j_i},Z_{l^j_i},\x_i)}}
\]
plus a quantity that behaves like $\O{s\sqrt\frac{\log\frac{1}{\delta}}{n}}$. We analyze the expected supr\=emus deviation term below:
\begin{eqnarray*}
&& \EE{(\x_i,\y_i,l^j_i)}{\usup{Z\in\Z}\bc{s\EE{(\x,\y,l)}{\ell_l(\y^l,Z_l,\x)} - \frac{1}{n}\sum_{i=1}^{n}\sum_{j=1}^s\ell(\y_i^{l^j_i},Z_{l^j_i},\x_i)}}\\
&=& \EE{(\x_i,\y_i,l^j_i)}{\usup{Z\in\Z}\bc{\sum_{j=1}^s\br{\EE{(\x,\y,l)}{\ell_l(\y^l,Z_l,\x)} - \frac{1}{n}\sum_{i=1}^{n}\ell(\y_i^{l^j_i},Z_{l^j_i},\x_i)}}}\\
&\leq& \sum_{j=1}^s\EE{(\x_i,\y_i,l^j_i)}{\usup{Z\in\Z}\bc{{\EE{(\x,\y,l)}{\ell_l(\y^l,Z_l,\x)} - \frac{1}{n}\sum_{i=1}^{n}\ell(\y_i^{l^j_i},Z_{l^j_i},\x_i)}}}\\
&\leq& \sum_{j=1}^s\frac{4C}{n}\EE{X,\vecl^j,\veceps}{\usup{Z\in\Z}{\ip{Z}{X^{\vecl^j}_\veceps}}} = \frac{4Cs}{n}\EE{X,\vecl,\veceps}{\usup{Z\in\Z}{\ip{Z}{X^{\vecl}_\veceps}}} = 4Cs\RR_{n}\br{\Z}
\end{eqnarray*}
and thus, it just suffices to prove bounds for the case where a single label is observed per point.

\subsection{Step 3: Estimating the Rademacher Average}
We will now bound the following quantity:
\[
\RR_n(\Z) = \frac{1}{n}\EE{X,\vecl,\veceps}{\usup{Z \in \Z}{\ip{Z}{X^{\vecl}_\veceps}}}
\]
where $X^{\vecl}_\veceps$ is as defined above. Approaches to bounding such Rademacher average terms usually resort to Martingale techniques \cite{KST09} or use of tools from convex analysis \cite{KST12} and decompose the Rademacher average term. However, such decompositions shall yield suboptimal results in our case. Our proposed approach will, instead involve an application of H\"older's inequality followed by an application from results from random matrix theory to bound the spectral norm of a random matrix.

For simplicity of notation, for any $l \in [L]$, we denote $V_l = \sum_{i \in I_l}\epsilon_ix_i$ and $V := X^\vecl_\veceps = [V_1\ V_2 \ldots V_L]$. Also, for any $l \in [L]$, let $n_l = \abs{I_l}$ denote the number of training points for which values of the $l\th$ label was observed i.e. $n_l = \sum_{i=1}^{n}\ind_{l_i=l}$.

\subsubsection{Distribution Independent Bound}
We apply H\"older's inequality to get the following result:	
\[
\frac{1}{n}\EE{X,\vecl,\veceps}{\usup{Z \in \Z}{\ip{Z}{X^{\vecl}_\veceps}}} \leq \frac{1}{n}\EE{X,\vecl,\veceps}{\usup{Z \in \Z}{\norm{Z}_{\tr}\norm{X^{\vecl}_\veceps}_F}} \leq \frac{1}{n}\EE{X,\vecl,\veceps}{\lambda\norm{X^{\vecl}_\veceps}_2} \leq \frac{\lambda}{n}\sqrt{\EE{X,\vecl,\veceps}{\norm{X^{\vecl}_\veceps}^2_2}}
\]
Then the following bound can be derived in a straightforward manner:
\begin{eqnarray*}
\EE{X,\vecl,\veceps}{\norm{X^\vecl_\veceps}_2^2} &\leq& \EE{X,\vecl,\veceps}{\norm{X^\vecl_\veceps}_F^2} = \EE{X,\vecl,\veceps}{\sum_{l=1}^L\norm{V_l}_2^2} = \EE{X,\vecl,\veceps}{\sum_{l=1}^L\norm{\sum_{i \in I_l}\epsilon_i\x_i}_2^2}\\
							 &=& \EE{X,\vecl,\veceps}{\sum_{l=1}^L\sum_{i \in I_l}\norm{\x_i}_2^2 + \sum_{i \neq j \in I_l}\epsilon_i\epsilon_j\ip{\x_i}{\x_j}}\\
							 &\leq& \EE{\vecl}{\sum_{l=1}^Ln_l\E{\norm{\x}_2^2}} \leq \EE{\vecl}{\sum_{l=1}^Ln_l} = n
\end{eqnarray*}
where we have assumed, without loss of generality that $\EE{\x \sim \D}{\norm{\x}_2^2} \leq 1$. This proves
\[
\RR_n(\Z) \leq \frac{\lambda}{\sqrt{n}},
\]
which establishes Theorem~\ref{thm:genbound-prelim}. Note that the same analysis holds if $Z$ is Frobenius norm regularized since we can apply the H\"older's inequality for Frobenius norm instead and still get the same Rademacher average bound.

\subsubsection{Tighter Bounds for Trace Norm Regularization}
Notice that in the above analysis, we did not exploit the fact that the top singular value of the matrix $X^\vecl_\veceps$ could be much smaller than its Frobenius norm. However, there exist distributions where trace norm regularization enjoys better performance guarantees over Frobenius norm regularization. In order to better present our bounds, we model the data distribution $\D$ on $\X$ (or rather its marginal) more carefully. Let $X := \E{\x\x^\top}$ and suppose the distribution $\D$ satisfies the following conditions:
\begin{enumerate}
	\item The top singular value of $X$ is $\norm{X}_2 = \sigma_1$
	\item The matrix $X$ has trace $\trr{X} = \Sigma$
	\item The distribution on $\X$ is sub-Gaussian i.e. for some $\eta > 0$, we have, for all $\v \in \R^d$,
	\[
	\E{\exp\br{x^\top \v}} \leq \exp\br{\norm{\v}_2^2\eta^2/2}
	\]
\end{enumerate}
In order to be consistent with previous results, we shall normalize the vectors $x$ so that they are unit-norm \emph{on expectation}. Since $\E{\norm{x}_2^2} = \trr{X} = \Sigma$, we wish to bound the Rademacher average as
\[
\RR_{n}\br{\Z} \leq \frac{1}{n\sqrt \Sigma}\EE{X,\vecl,\veceps}{\usup{Z \in \Z}{\ip{Z}{X^{\vecl}_\veceps}}}
\]
In this case, it is possible to apply the H\"older's inequality as
\[
\frac{1}{n\sqrt \Sigma}\EE{X,\vecl,\veceps}{\usup{Z \in \Z}{\ip{Z}{X^{\vecl}_\veceps}}} \leq \frac{1}{n\sqrt \Sigma}\EE{X,\vecl,\veceps}{\usup{Z \in \Z}{\norm{Z}_{\tr}\norm{X^{\vecl}_\veceps}_2}} \leq \frac{1}{n\sqrt \Sigma}\EE{X,\vecl,\veceps}{\lambda\norm{X^{\vecl}_\veceps}_2} \leq \frac{\lambda}{n\sqrt \Sigma}\sqrt{\EE{X,\vecl,\veceps}{\norm{X^{\vecl}_\veceps}^2_2}}
\]
Thus, in order to bound $\RR_n(\Z)$, it suffices to bound $\EE{X,\vecl,\veceps}{\norm{X^\vecl_\veceps}_2^2}$. In this case, since our object of interest is the spectral norm of the matrix $X^\vecl_\veceps$, we expect to get much better guarantees, for instance, in case the training points $x \in \X$ are being sampled from some (near) isotropic distribution. We note that Frobenius norm regularization will not be able to gain any advantage in these situations since it would involve the Frobenius norm of the matrix $X^\vecl_\veceps$ (as shown in the previous subsubsection) and thus, cannot exploit the fact that the spectral norm of this matrix is much smaller than its Frobenius norm.

\subsection{Step 4: Calculating the Spectral norm of a Random Matrix}
To bound $\EE{X,\vecl,\veceps}{\norm{X^\vecl_\veceps}_2^2}$, we first make some simplifications (we will take care of the normalizations later). For any $l \in [L]$, let the probability of the value for label $l$ being observed be $p_l \in (0,1]$ such that $\sum_lp_l = 1$. Also let $P = \underset{l \in [L]}{\max}\ p_l$ and $p = \underset{l \in [L]}{\min}\ p_t$. Call the event $\event_{\max}$ as the event when $n_l \leq 2P\cdot n$ for all $l \in [L]$ i.e. every label will have at most $2P\cdot n$ training points for which its value is seen. The following result shows that this is a high probability event:
\begin{lem}
\label{per-label-obs}
For any $\delta > 0$, if $n \geq {\frac{1}{2p^2}\log\frac{L}{\delta}}$, then with probability $1 - \delta$, we have
\[
\Pr{\event_{\max}} \geq 1 - \delta
\]
\end{lem}
\begin{proof}
For any $l \in [L]$, an application of Chernoff's bound for Boolean random variables tells us that with probability at least $1 - \exp\br{{-2np_l^2}}$, we have $n_l \leq 2p_l\cdot n \leq 2P\cdot n$. Taking a union bound and using $p_l \geq p$ finishes the proof.
\end{proof}
Conditioning on the event $\event_{\max}$ shall allow us to get a control over the spectral norm of the matrix $X^\vecl_\veceps$ by getting a bound on the sub-Gaussian norm of the individual columns of $X^\vecl_\veceps$. We show below, that conditioning on this event does not affect the Rademacher average calculations. A simple calculation shows that $\EE{X,\veceps}{\left.\norm{X^\vecl_\veceps}_2^2\right|\vecl} \leq n\Sigma$. If we have $n > \frac{1}{2p^{2}}\log \frac{L\Sigma}{Pd(\eta^2+\sigma_1)}$, we have $\Pr{\neg\event_{\max}} < {\frac{Pd(\eta^2 + \sigma_1)}{\Sigma}}$. This gives us the following bound:

\begin{eqnarray*}
\EE{X,\vecl,\veceps}{\norm{X^\vecl_\veceps}_2^2} &=& \EE{X,\veceps}{\left.\norm{X^\vecl_\veceps}_2^2\right|\event_{\max}}\Pr{\event_{\min}} + \EE{X,\veceps}{\left.\norm{X^\vecl_\veceps}_2^2\right|\neg\event_{\min}}\br{1 - \Pr{\event_{\max}}}\\
												 &=& \EE{X,\veceps}{\left.\norm{X^\vecl_\veceps}_2^2\right|\event_{\max}}(1 - \delta) + \EE{X,\veceps}{\left.\norm{X^\vecl_\veceps}_2^2\right|\neg\event_{\max}}\delta\\
												 &\leq& \EE{X,\veceps}{\left.\norm{X^\vecl_\veceps}_2^2\right|\event_{\max}} + n\Sigma\br{\frac{Pd(\eta^2 + \sigma_1)}{\Sigma}}\\
												 &\leq& \O{\EE{X,\veceps}{\left.\norm{X^\vecl_\veceps}_2^2\right|\event_{\max}}}
\end{eqnarray*}
where the last step follows since our subsequent calculations will show that $\EE{X,\veceps}{\left.\norm{X^\vecl_\veceps}_2^2\right|\event_{\max}} = \O{nPd(\eta^2 +\sigma_1)}$. Thus, it suffices to bound $\EE{X,\veceps}{\left.\norm{X^\vecl_\veceps}_2^2\right|\event_{\max}} = \EE{X,\veceps}{\left.\norm{V}_2^2\right|\event_{\max}}$. For sake of brevity we will omit the conditioning term from now on.

For simplicity let $A_l = \frac{V_l}{c}$ where $c = \eta\cdot\sqrt{2P\cdot n}$ and $A = \bs{A_1 A_2 \ldots A_L}$. Thus
\[
\EE{X,\vecl,\veceps}{\norm{X^\vecl_\veceps}_2^2} = c^2\cdot\EE{X,\vecl,\veceps}{\norm{A}_2^2}
\]
We first bound the sub-Gaussian norm of the column vectors $A_l$. For any vector $\v \in \R^d$, we have:
\begin{eqnarray*}
\E{\exp\br{A_l^\top \v}} &=& \E{\exp\br{\frac{1}{c}\sum_{i \in I_l}\epsilon_i\ip{\x_i}{\v}}}\\
						&=& \br{\E{\exp\br{\ip{\x}{\frac{1}{c}\epsilon \v}}}}^{n_l}\\
						&\leq& \br{{\exp\br{\norm{\frac{1}{c}\epsilon \v}_2^2\eta ^2/2}}}^{n_l}\\
						&=& {\exp\br{\frac{n_l}{2\eta^2P\cdot n}\norm{\v}_2^2\eta ^2/2}}\\
						&\leq& \exp\br{\norm{\v}_2^2/2}
\end{eqnarray*}
where, in the second step, we have used the fact that $\x_i,\x_j$ and $\epsilon_i,\epsilon_j$ are independent for $i \neq j$, in the third step we have used the sub-Gaussian properties of $\x$ and in the fourth step, we have use the fact that the event $\event_{\max}$ holds. This shows us that the sub-Gaussian norm of the column vector $A_l$ is bounded i.e. $\norm{A_l}_{\psi_2} \leq 1$.

We now proceed to bound $\EE{X,\veceps}{\norm{A}_2^2} = \EE{X,\veceps}{\norm{A^\top}_2^2}$. Our proof proceeds by an application of a Bernstein-type inequality followed by a covering number argument and finishing off by bounding the expectation in terms of the cumulative distribution function. The first two parts of the proof proceed on the lines of the proof of Theorem 5.39 in \cite{V12} For any fixed vector $\v \in \S^{d-1}$, the set of unit norm vectors in $d$ dimensions, we have:
\[
{\norm{A\v}_2^2} = \sum_{l=1}^L{\ip{A_l}{\v}^2} =: \sum_{l=1}^LZ_l^2
\]
Now observe that conditioned on $\vecl$, $I_t \cap I_{t'} = \phi$ if $t \neq t'$ and thus, conditioned on $\vecl$, the variables $Z_t, Z_{t'}$ are independent for $t \neq t'$. This will allow us to apply the following Bernstein-type inequality
\begin{thm}[\cite{V12}, Corollary 5.17]
\label{thm:bernstein-subexp}
Let $X_1, \ldots, X_N$ be independent centered sub-exponential variables with bounded sub-exponential norm i.e. for all $i$, we have $\norm{X_i}_{\psi_1} \leq B$ for some $B > 0$. Then for some absolute constant $c_1 > 0$, we have for any $\epsilon > 0$,
\[
\Pr{\sum_{i=1}^NX_i \geq \epsilon N} \leq \exp\br{-c_1\min\bc{\frac{\epsilon^2}{B^2},\frac{\epsilon}{B}}N}.
\]
\end{thm}
To apply the above result, we will first bound expectation of the random variables $Z_l^2$.
\[
\E{Z_l^2} = \E{\ip{A_l}{\v}^2} = \E{\br{\frac{1}{c}\sum_{i \in I_l}\epsilon_i\ip{\x_i}{\v}}^2} = \frac{n_l}{c^2}\E{\ip{\x}{\v}^2} \leq \frac{n_l\sigma_1}{c^2} \leq \frac{\sigma_1}{\eta^2}
\]
where the fourth inequality follows from definition of the top singular norm $\sigma_1$ of $X := \E{xx^\top}$ and the last inequality follows from the event $\event_{\max}$. The above calculation gives us a bound on the expectation of $Z_l^2$ which will be used to center it. Since we have already established $\norm{A_l}_{\psi_2} \leq 1$, we automatically get $\norm{Z_l}_{\psi_2} \leq 1$.  Using standard inequalities between the sub-exponential norm $\norm{\cdot}_{\psi_1}$ and the sub-Gaussian norm $\norm{\cdot}_{\psi_2}$ of random variables \citep[for instance, see][Lemma 5.14]{V12} we also have
\[
\norm{Z_l^2 - \E{Z_l^2}}_{\psi_1} \leq 2\norm{Z_l^2}_{\psi_1} \leq 4\norm{Z_l}_{\psi_2}^2 \leq 4.
\]
Applying Theorem~\ref{thm:bernstein-subexp} to the variables $X_l = Z_l^2 - \E{Z_l^2}$, we get
\[
\Pr{\sum_{l=1}^LZ_l^2 - L\frac{\sigma_1}{\eta^2} \geq \epsilon L} \leq \exp\br{-c_1L\min\bc{\epsilon^2,\epsilon}}
\]
where $c_1 > 0$ is an absolute constant. Thus with probability at least $1 - \exp\br{-c_1L\min\bc{\epsilon^2,\epsilon}}$, for a fixed vector $\v \in \S^{d-1}$, we have the inequality
\[
{\norm{A\v}_2^2} \leq \br{\frac{\sigma_1}{\eta^2} + \epsilon}L
\]
Applying a union bound over a $\frac{1}{4}$-net $\Nr_{1/4}$ over $\S^{d-1}$ (which can be of size at most $9^d$), we get that with probability at most $1 - 9^d\exp\br{-c_1L\min\bc{\epsilon^2,\epsilon}}$, we have the above inequality for every vector $\v \in \Nr_{1/4}$ as well. We note that this implies a bound on the spectral norm of the matrix $A$ \citep[see][Lemma 5.4]{V12} and get the following bound
\[
\norm{A}_2^2 \leq 2\br{\frac{\sigma_1}{\eta^2} + \epsilon}L
\]
Put $\epsilon = c_2\cdot\frac{d}{L}+\frac{\epsilon'}{L}$ where $c_2 = \max\bc{1,\frac{\ln 9}{c_1}}$ and suppose $d \geq L$. Since $c_2 \geq 1$, we have $\epsilon \geq 1$ which gives $\min\bc{\epsilon,\epsilon^2} = \epsilon$. This gives us with probability at least $1 - \exp\br{-c_1\epsilon'}$,
\[
\norm{A}_2^2 \leq 2\br{L\frac{\sigma_1}{\eta^2} + c_2d + \epsilon'}
\]
Consider the random variable $Y = \frac{\norm{A}_2^2}{2}- L\frac{\sigma_1}{\eta^2} - c_2d$. Then we have $\Pr{Y >\epsilon} \leq \exp\br{-c_1\epsilon}$. Thus we have
\[
\E{Y} = \int_0^\infty\Pr{Y > \epsilon}d\epsilon \leq \int_0^\infty\exp\br{-c_1\epsilon}d \epsilon = \frac{1}{c_1}
\]
This gives us
\[
\E{{\norm{A}_2^2}} \leq 2\br{L\frac{\sigma_1}{\eta^2} + c_2d + \frac{1}{c_1}}
\]
and consequently,
\[
\EE{X,\vecl,\veceps}{\norm{X^\vecl_\veceps}_2^2} = c^2 \cdot\EE{X,\vecl,\veceps}{\norm{A}_2^2} \leq 4\eta^2P\cdot n\br{L\frac{\sigma_1}{\eta^2} + c_2d + \frac{1}{c_1}} \leq \O{n\eta^2P\br{d + L\frac{\sigma_1}{\eta^2}}} \leq \O{nPd(\eta^2 + \sigma_1)}
\]
where the last step holds when $d \geq L$. Thus, we are able to bound the Rademacher averages, for some absolute constant $c_3$ as
\[
\RR_{n}\br{\Z} \leq \frac{\lambda}{n\sqrt\Sigma}\sqrt{\norm{\EE{X,\vecl,\veceps}{\norm{X^\vecl_\veceps}_2^2}}} \leq c_3\lambda\sqrt\frac{Pd(\eta^2+\sigma_1)}{n\Sigma},
\]
which allows us to make the following claim:
\begin{thm}
\label{thm:genbound-general}
Suppose we learn a predictor using the trace norm regularized formulation $\hat Z = \underset{\norm{Z}_{\tr} \leq \lambda}{\arg\inf}\ \hat\L(Z)$ over a set of $n$ training points. Further suppose that, for any $l \in [L]$, the probability of observing the value of label $l$ is given by $p_l$ and let $P = \underset{l \in [L]}{\max}\ p_l$. Then with probability at least $1 - \delta$, we have
\[
\L(\hat Z) \leq \underset{\norm{Z}_{\tr} \leq \lambda}{\arg\inf}\ \L(Z) + \O{s\lambda\sqrt\frac{dP(\eta^2+\sigma_1)}{n\Sigma}} + \O{s\sqrt\frac{\log\frac{1}{\delta}}{{n}}},
\]
where the terms $\eta,\sigma_1,\Sigma$ are defined by the data distribution as before.
\end{thm}
Essentially, the above result indicates that if some label is observed too often, as would be the case when $P = \Om{1}$, we get no benefit from trace norm regularization since this is akin to a situation with fully observed labels. However, if the distribution on the labels is close to uniform i.e. $P = \O{\frac{1}{L}}$, the above calculation lets us bound the Rademacher average, and consequently, the excess risk as
\[
\RR_{n}\br{\Z} \leq c_3\lambda\sqrt\frac{d(\eta^2+\sigma_1)}{nL\Sigma},
\]
thus proving the first part of Theorem~\ref{thm:genbound-main}. 

We now notice that 
However, in case our data distribution is near isotropic, i.e. $\Sigma \gg \sigma_1$, then this result gives us superior bounds. For instance, if the data points are generated from a standard normal distribution, then we have $\sigma_1 = 1$, $\Sigma = d$ and $\eta = 1$ using which we can bound the Rademacher average term as
\[
\RR_{n}\br{\Z} \leq c_3\lambda\sqrt\frac{2}{nL},
\]
which gives us the second part of Theorem~\ref{thm:genbound-main}.

\section{Lower Bounds for Uniform Convergence-based Proofs}
In this section, we show that our analysis for Theorems~\ref{thm:genbound-prelim} and \ref{thm:genbound-main} are essentially tight. In particular, we show for each case, a data distribution such that the deviation of the empirical losses from the population risks is, up to a constant factor, the same as predicted by the results. We state these lower bounds in two separate subsections below:

\subsection{Lower Bound for Trace Norm Regularization}
\label{appsec:lower-bound-trace}
In this section we shall show that for general distribution, Theorem~\ref{thm:genbound-prelim} is tight. Recall that Theorem~\ref{thm:genbound-prelim} predicts that for a predictor $\hat Z$ learned using a trace norm regularized formulation satisfies, with constant probability (i.e. $\delta = \Om{1}$),
\[
\L(\hat Z) \leq \hat\L(\hat Z) + \O{\lambda\sqrt\frac{1}{n}},
\]
where, for simplicity as well as w.l.o.g., we have assumed $s = 1$. We shall show that this result is tight by demonstrating the following lower bound:
\begin{claim}
\label{claim:lower-bound-trace}
There exists a data-label distribution and a loss function such that the empirical risk minimizer learned as $\hat Z = \underset{\norm{Z}_{\tr} \leq \lambda}{\arg\inf}\ \hat\L(Z)$ has, with constant probability, its population risk lower bounded by 
\[
\L(\hat Z) \geq \hat\L(\hat Z) + \Om{\lambda \sqrt\frac{1}{n}},
\]
\end{claim}
thus establishing the tightness claim. Our proof will essentially demonstrate this by considering a non-isotropic data distribution (since, for isotropic distributions, Theorem~\ref{thm:genbound-main} shows that a tighter upper bound is actually possible). For simplicity, and w.l.o.g., we will prove the result for $\lambda = 1$. Let $\vecmu \in \R^d$ be a fixed unit vector and consider the following data distribution
\[
\x_i = \zeta_i\vecmu,
\]
where $\zeta_i$ are independent Rademacher variables and a trivial label distribution
\[
\y_i = \ind,
\]
where $\ind \in \R^L$ is the all-ones vector. Note that the data distribution satisfies $\E{\norm{\x}_2^2} = 1$ and thus, satisfies the assumptions of Theorem~\ref{thm:genbound-prelim}. Let $\omega_i^l = 1$ iff the label $l$ is observed for the $i\th$ training point. Note that for any $i$, we have $\sum_{l=1}^L\omega_i^l = 1$ and that for any $l \in [L]$, $\omega_i^l = 1$ with probability $1/L$. Also consider the following loss function
\[
\ell(\y^{l},f^{l}(\x;Z)) = \ip{Z_l}{\y^{l}\x}
\]
Let
\[
\hat Z = \underset{\norm{Z}_{\tr} \leq 1}{\arg\inf}\ \hat\L(Z) = \underset{\norm{Z}_{\tr} \leq 1}{\arg\inf}\ \frac{1}{n}\ip{Z}{\vecmu\vv^\top}
\]
where $\vv$ is the vector
\[
\vv = \bs{\sum_{i=1}^n\zeta_i\omega^1_i\ \sum_{i=1}^n\zeta_i\omega^2_i\ \ldots\ \sum_{i=1}^n\zeta_i\omega^L_i}
\]
Clearly, since $x$ is a centered distribution and $\ell$ is a linear loss function, $\L(\hat Z) = 0$. However, by H\"older's inequality, we also have
\[
\hat Z = -\frac{\vecmu\vv^\top}{\norm{\vv}_2},
\]
and thus, $\hat\L(\hat Z) = -\frac{1}{n}\norm{\vv}_2$ since $\norm{\vecmu}_2 = 1$. The following lemma shows that with constant probability, $\norm{\vv}_2 \geq \sqrt{n/2}$ which shows that $\L(\hat Z) \geq \hat\L(\hat Z) + \Om{\sqrt\frac{1}{n}}$, thus proving the lower bound.

\begin{lem}
With probability at least $3/4$, we have $\norm{\vv}_2^2 \geq n/2$.
\end{lem}
\begin{proof}
We have
\begin{eqnarray*}
\norm{\vv}_2^2 &=& \sum_{l=1}^L\br{\sum_{i=1}^n\zeta_i\omega^l_i}^2 = \sum_{l=1}^L\sum_{i=1}^n\omega^l_i + \sum_{l=1}^L\sum_{i\neq j}\zeta_i\omega^l_i\zeta_j\omega^l_j\\
							 &=& n + \sum_{i\neq j}\zeta_i\zeta_j\ip{\vecom_i}{\vecom_j} = n + W,
\end{eqnarray*}
where $\vecom_i = [\omega_i^1,\omega_i^2,\ldots,\omega_i^L]$. Now clearly $\E{W} = 0$ and as the following calculation shows, $\E{W^2} \leq 2n^2/L$ which, by an application of Tchebysheff's inequality, gives us, for $L > 32$, with probability at least $3/4$, $\abs{W} \leq n/2$ and consequently $\norm{\vv}_2^2 \geq n/2$. We give an estimation of the variance of $Z$ below.
\begin{eqnarray*}
\E{W^2} &=& \E{\sum_{i_1\neq j_1, i_2 \neq j_2} \zeta_{i_1}\zeta_{j_1}\ip{\vecom_{i_1}}{\vecom_{j_1}}\zeta_{i_2}\zeta_{j_2}\ip{\vecom_{i_2}}{\vecom_{j_2}}}\\
		&=& 2\E{\sum_{i \neq j} \ip{\vecom_{i}}{\vecom_{j}}^2} = 2n(n-1)\E{\ip{\vecom_1}{\vecom_2}} \leq \frac{2n^2}{L},
\end{eqnarray*}
where we have used the fact that ${\ip{\vecom_i}{\vecom_j}}^2 = \ip{\vecom_i}{\vecom_j}$ since $\ip{\vecom_i}{\vecom_j} = 0 \text{ or } 1$, and that $\E{\ip{\vecom_1}{\vecom_2}} = \frac{1}{L}$ since that is the probability of the same label getting observed for $\x_1$ and $\x_2$.
\end{proof}

\subsection{Lower Bound for Frobenius Norm Regularization}
\label{appsec:lower-bound-frob}
In this section, we shall prove that even for isotropic distributions, Frobenius norm regularization cannot offer $\O{\frac{1}{\sqrt{nL}}}$-style bounds as offered by trace norm regularization.
\begin{claim}
\label{claim:lower-bound-frob}
There exists an isotropic, sub-Gaussian data distribution and a loss function such that the empirical risk minimizer learned as $\hat Z = \underset{\norm{Z}_F \leq \lambda}{\arg\inf}\ \hat\L(Z)$ has, with constant probability, its population risk lower bounded by 
\[
\L(\hat Z) \geq \hat\L(\hat Z) + \Om{\lambda\sqrt\frac{1}{n}},
\]
whereas an empirical risk minimizer learned as $\hat Z = \underset{\norm{Z}_{\tr} \leq \lambda}{\arg\inf}\ \hat\L(Z)$ over the same distribution has, with probability at least $1 - \delta$, its population risk bounded by 
\[
\L(\hat Z) \leq \hat\L(\hat Z) + \O{\lambda\sqrt\frac{1}{nL}} + \O{\sqrt\frac{\log\frac{1}{\delta}}{n}}.
\]
\end{claim}
We shall again prove this result for $\lambda = 1$. We shall retain the distribution over labels as well as the loss function from our previous discussion in Appendix~\ref{appsec:lower-bound-trace}. We shall also reuse $\omega_i^l$ to denote the label observation pattern. We shall however use Rademacher vectors to define the data distribution i.e. each of the $d$ coordinates of the vector $\x$ obeys the law
\[
r \sim \frac{1}{2}(\ind_{\bc{r = 1}} + \ind_{\bc{r = -1}}).
\]
Thus we sample $\x_i$ as
\[
\x_i = \frac{1}{\sqrt{d}}\bs{r_i^1,\ r_i^2,\ \ldots\ ,r_i^d},
\]
where each coordinate is independently sampled. We now show that this distribution satisfies the assumptions of Theorem~\ref{thm:genbound-main}. We have $\E{\x\x^\top} = \frac{1}{d}\cdot\mathbb{I}$ where $\mathbb{I}$ is the $d \times d$ identity matrix. Thus $\sigma_1 = \frac{1}{d}$ and $\Sigma = 1$. We also have, for any $\v \in \R^d$,
\begin{eqnarray*}
\E{\exp\br{\x^\top\v}} &=& \E{\exp\br{\sum_{j=1}^d\x^j\v^j}} = \prod_{j=1}^d\E{\exp\br{\x^j\v^j}}\\
					   &=& \prod_{j=1}^d\frac{1}{2}\br{\exp\br{\frac{1}{\sqrt d}\v^j} + \exp\br{-\frac{1}{\sqrt d}\v^j}}\\
					   &=& \prod_{j=1}^d\cosh\br{\frac{1}{\sqrt d}\v^j} \leq \prod_{j=1}^d\exp\br{\frac{1}{d}(\v^j)^2}\\
					   &=& \exp\br{\sum_{j=1}^d\frac{1}{d}(\v^j)^2} = \exp\br{\frac{1}{d}\norm{\v}_2^2},
\end{eqnarray*}
where the second equality uses the independence of the coordinates of $\x$. Thus we have $\eta^2 = \frac{2}{d}$. Thus, this distribution fulfills all the preconditions of Theorem~\ref{thm:genbound-main}. Note that had trace norm regularization been applied, then by applying Theorem~\ref{thm:genbound-main}, we would have gotten an excess error of
\[
\O{\sqrt\frac{d(\eta^2 + \sigma_1)}{nL\Sigma}} = \O{\sqrt\frac{d(2/d + 1/d)}{nL\cdot 1}} = \O{\sqrt\frac{1}{nL}}
\]
whereas, as the calculation given below shows, Frobenius norm regularization cannot guarantee an excess risk better than $\O{\sqrt\frac{1}{n}}$. Suppose we do perform Frobenius norm regularization in this case. Then we have
\[
\hat Z = \underset{\norm{Z}_{F} \leq 1}{\arg\inf}\ \hat\L(Z) = \underset{\norm{Z}_{F} \leq 1}{\arg\inf}\ \frac{1}{n}\ip{Z}{X},
\]
where $X$ is the matrix
\[
X = \bs{\sum_{i=1}^L\omega_i^1\x_i\ \sum_{i=1}^L\omega_i^2\x_i\ \ldots\ \sum_{i=1}^L\omega_i^L\x_i}.
\]
As before, $\L(\hat Z) = 0$ since the data distribution is centered and the loss function is linear. By a similar application of H\"older's inequality, we can also get
\[
\hat Z = -\frac{X}{\norm{X}_F},
\]
and thus, $\hat\L(\hat Z) = -\frac{1}{n}\norm{X}_F$. The following lemma shows that with constant probability, $\norm{X}_F \geq \sqrt{n/2}$ which shows that $\L(\hat Z) \geq \hat\L(\hat Z) + \Om{\sqrt\frac{1}{n}}$, thus proving the claimed inability of Frobenius norm regularization to give $\O{\frac{1}{\sqrt{nL}}}$-style bounds even for isotropic distributions.

\begin{lem}
With probability at least $3/4$, we have $\norm{X}_F^2 \geq n/2$.
\end{lem}
\begin{proof}
We have
\begin{eqnarray*}
\norm{X}_F^2 &=& \sum_{l=1}^L\norm{\sum_{i=1}^n\omega^l_i\x_i}_2^2 = \sum_{l=1}^L\sum_{i=1}^n\omega^l_i\norm{\x_i}_2^2 + \sum_{l=1}^L\sum_{i\neq j}\omega^l_i\omega^l_j\ip{\x_i}{\x_j}\\
			 &=& \sum_{i=1}^n \norm{\x_i}_2^2 + \sum_{i\neq j}\ip{\x_i}{\x_j}\ip{\vecom_i}{\vecom_j} = n + W
\end{eqnarray*}
where as before, $\vecom_i = [\omega_i^1,\omega_i^2,\ldots,\omega_i^L]$. We will, in the sequel prove that $\abs{W} \leq n/2$, thus establishing the claim. Clearly $\E{W} = 0$ and as the following calculation shows, $\E{W^2} \leq 2n^2/Ld$ which, by an application of Tchebysheff's inequality, gives us, for $Ld > 32$, with probability at least $3/4$, $\abs{W} \leq n/2$ and consequently $\norm{X}_F^2 \geq n/2$. We give an estimation of the variance of $W$ below.
\begin{eqnarray*}
\E{W^2} &=& \E{\sum_{i_1\neq j_1, i_2 \neq j_2} \ip{\x_{i_1}}{\x_{j_1}}\ip{\vecom_{i_1}}{\vecom_{j_1}}\ip{\x_{i_2}}{\x_{j_2}}\ip{\vecom_{i_2}}{\vecom_{j_2}}}\\
		&=& 2\E{\sum_{i \neq j} \ip{\x_i}{\x_j}^2\ip{\vecom_{i}}{\vecom_{j}}^2} = 2n(n-1)\E{\ip{\x_1}{\x_2}^2\ip{\vecom_1}{\vecom_2}}\\
		&=& 2n(n-1)\E{\ip{\x_1}{\x_2}^2}\E{\ip{\vecom_1}{\vecom_2}} \leq \frac{2n^2}{Ld},
\end{eqnarray*}
where we have used the fact that data points and label patterns are sampled independently.
\end{proof}


\section{More Experimental Results}
\label{app:exp}

\subsection{Speedup Results Due to Multi-core Computation}
\label{app:exp-speedup}
\begin{figure}[H]
  \begin{tabular}{cc}
   \subfloat[$\nabla g(\w)$]{
   \label{fig:grad_speedup}
   \includegraphics[width=0.5\linewidth]{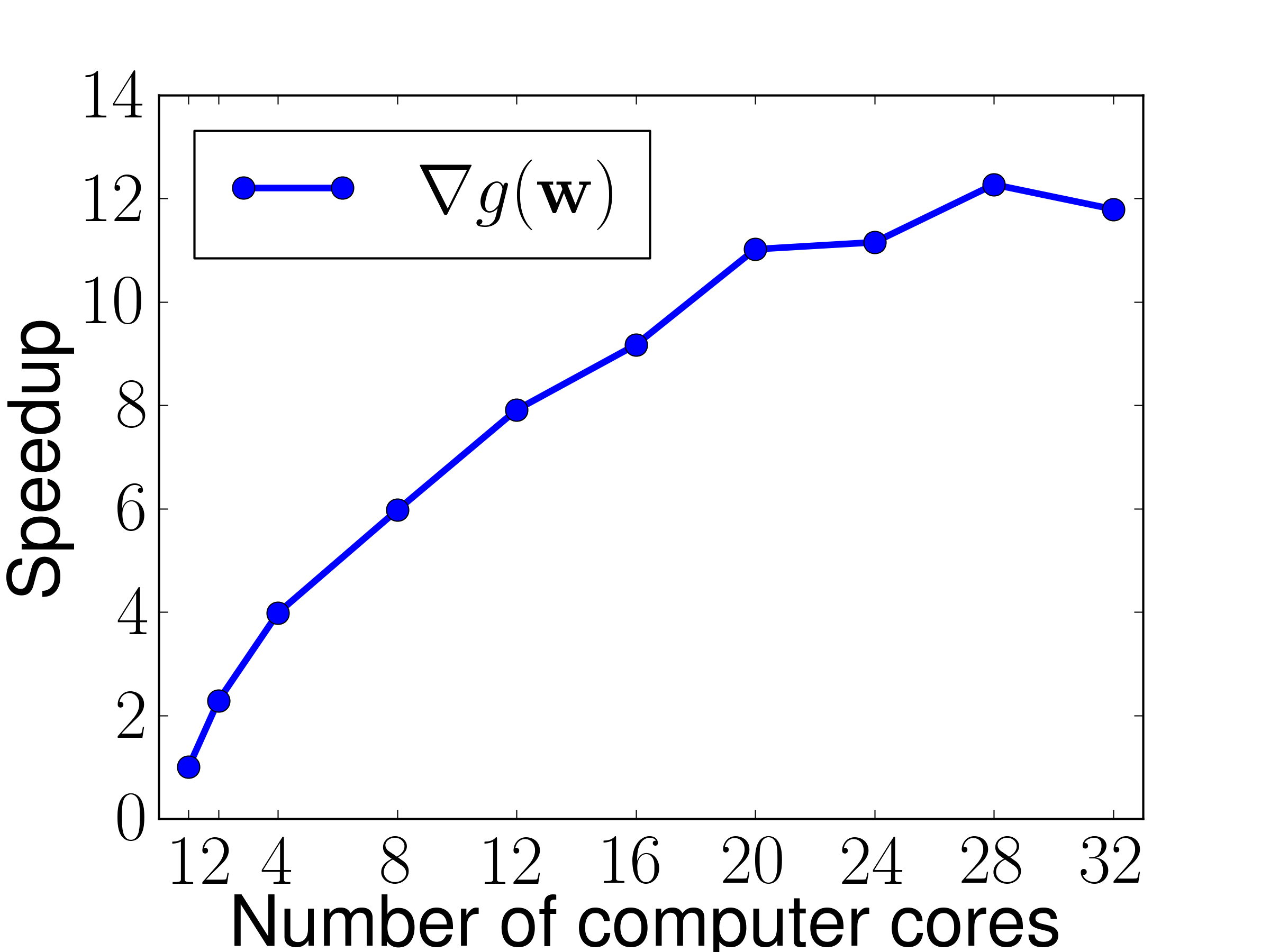}
   }   &
   \subfloat[$\nabla ^2 g(\w) \s$]{
   \label{fig:Hv_speedup}
   \includegraphics[width=0.5\linewidth]{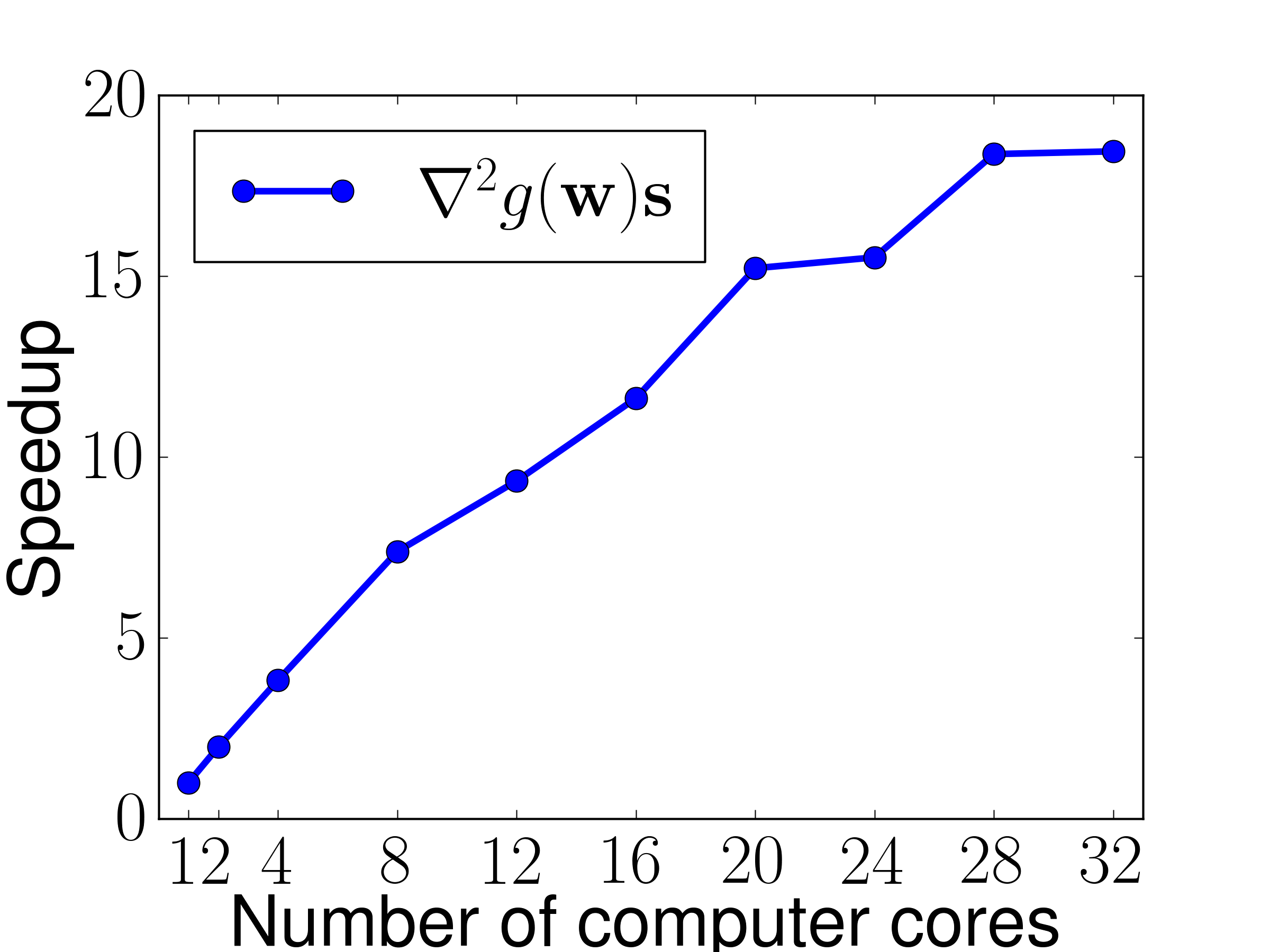}
   } 
  \end{tabular}
  \caption{Speedup results for our proposed fast gradient calculation and 
  Hessian-vector multiplication.}
  \label{fig:speedup}
\end{figure}
\subsection{Detailed Results with Full Labels}
\label{app:exp-full}
\begin{itemize}
  \item Table \ref{tab:Y-o-1-p-1} shows the top-1 accuracy results for the 
    case with fully observed labels.
  \item Table \ref{tab:Y-o-1-p-3} shows the top-3 accuracy results for the 
    case with fully observed labels.
  \item Table \ref{tab:Y-o-1-p-5} shows the top-5 accuracy results for the 
    case with fully observed labels.
  \item Table \ref{tab:Y-o-1-p-0} shows the Hamming loss results for the 
    case with fully observed labels.
  \item Table \ref{tab:Y-o-1-p--1} shows the average AUC results for the 
    case with fully observed labels.
\end{itemize}
\begin{table*}[h]
\caption{Comparison for dimensionality reductions approach on fully observed $Y$ with various rank. SQ for squared loss, LR for logistic loss, SH for squared hinge loss, and WAR for weighted approximated-rank loss}
\label{tab:Y-o-1-p-1}
\centering
\small
\begin{tabular}{lr|r|r|r|r|r|r}
	&	& \multicolumn{6}{@{}c@{}}{Top-$1$ Accuracy}\\
	&	& \multicolumn{3}{@{}c@{}|}{\MLR} & \BCS & \CPLST & \WSABIE \\
	&$k/L$ & SQ & LR & SH & SQ& SQ & WAR \\
\hline
\hline
\multirow{5}{*}{bibtex} 
&	$20\%$ & {\bf 58.33}& 46.20& 46.52& 41.43& 55.55& 48.51\\
&	$40\%$ & {\bf 60.99}& 50.78& 40.68& 54.63& 58.73& 52.37\\
&	$60\%$ & {\bf 61.99}& 51.37& 39.24& 57.53& 60.36& 51.45\\
&	$80\%$ & {\bf 63.38}& 52.64& 39.96& 59.76& 62.31& 53.04\\
&	$100\%$ & {\bf 63.94}& 53.76& 38.41& 60.24& 63.02& 53.24\\
\hline
\multirow{5}{*}{autofood} 
&	$20\%$ & 86.84& 84.21& {\bf 89.47}& 68.42& 52.63& 47.37\\
&	$40\%$ & {\bf 92.11}& 89.47& {\bf 92.11}& 28.95& 55.26& 86.84\\
&	$60\%$ & 73.68& {\bf 89.47}& 86.84& 71.05& 52.63& 65.79\\
&	$80\%$ & {\bf 94.74}& 89.47& 89.47& 81.58& 57.89& 78.95\\
&	$100\%$ & 81.58& {\bf 89.47}& 86.84& 84.21& 57.89& 60.53\\
\hline
\multirow{5}{*}{compphys} 
&	$20\%$ & 92.50& 87.50& {\bf 97.50}& 70.00& 52.50& 65.00\\
&	$40\%$ & {\bf 95.00}& 92.50& {\bf 95.00}& 65.00& 50.00& 47.50\\
&	$60\%$ & {\bf 95.00}& 92.50& {\bf 95.00}& 72.50& 47.50& 70.00\\
&	$80\%$ & 95.00& 87.50& {\bf 97.50}& 75.00& 50.00& 45.00\\
&	$100\%$ & 95.00& {\bf 97.50}& {\bf 97.50}& 67.50& 50.00& 52.50\\
\hline
\multirow{5}{*}{delicious} 
&	$20\%$ & {\bf 67.16}& 57.39& 61.07& 59.50& 66.53& 48.35\\
&	$40\%$ & {\bf 66.66}& 51.62& 56.20& 61.16& 66.25& 47.25\\
&	$60\%$ & {\bf 66.28}& 50.96& 51.59& 63.08& 66.22& 47.38\\
&	$80\%$ & {\bf 66.25}& 51.55& 49.11& 62.10& 66.22& 45.59\\
&	$100\%$ & {\bf 66.28}& 50.83& 46.53& 63.45& 66.22& 46.25\\
\end{tabular}
\end{table*}
\begin{table*}[h]
\caption{Comparison for dimensionality reductions approach on fully observed $Y$ with various rank. SQ for squared loss, LR for logistic loss, SH for squared hinge loss, and WAR for weighted approximated-rank loss}
\label{tab:Y-o-1-p-3}
\centering
\small
\begin{tabular}{lr|r|r|r|r|r|r}
	&	& \multicolumn{6}{@{}c@{}}{Top-$3$ Accuracy}\\
	&	& \multicolumn{3}{@{}c@{}|}{\MLR} & \BCS & \CPLST & \WSABIE \\
	&$k/L$ & SQ & LR & SH & SQ& SQ & WAR \\
\hline
\hline
\multirow{5}{*}{bibtex} 
&	$20\%$ & {\bf 34.16}& 25.65& 27.37& 21.74& 31.99& 28.77\\
&	$40\%$ & {\bf 36.53}& 28.20& 24.81& 28.95& 34.53& 30.05\\
&	$60\%$ & {\bf 38.00}& 28.68& 23.26& 32.25& 36.01& 31.11\\
&	$80\%$ & {\bf 38.58}& 29.42& 23.04& 34.09& 36.75& 31.21\\
&	$100\%$ & {\bf 38.41}& 30.25& 22.36& 34.87& 36.91& 31.24\\
\hline
\multirow{5}{*}{autofood} 
&	$20\%$ & {\bf 81.58}& 80.70& {\bf 81.58}& 53.51& 42.98& 66.67\\
&	$40\%$ & 76.32& {\bf 80.70}& 78.95& 50.88& 42.11& 70.18\\
&	$60\%$ & 70.18& 80.70& {\bf 81.58}& 64.91& 41.23& 60.53\\
&	$80\%$ & 80.70& 80.70& {\bf 85.09}& 73.68& 42.98& 72.81\\
&	$100\%$ & 75.44& 80.70& {\bf 82.46}& 65.79& 42.98& 64.04\\
\hline
\multirow{5}{*}{compphys} 
&	$20\%$ & {\bf 80.00}& {\bf 80.00}& {\bf 80.00}& 42.50& 40.83& 49.17\\
&	$40\%$ & {\bf 80.00}& 78.33& 79.17& 60.00& 37.50& 39.17\\
&	$60\%$ & {\bf 80.00}& {\bf 80.00}& {\bf 80.00}& 51.67& 39.17& 49.17\\
&	$80\%$ & 80.00& 78.33& {\bf 80.83}& 53.33& 39.17& 52.50\\
&	$100\%$ & 80.00& 79.17& {\bf 81.67}& 62.50& 39.17& 56.67\\
\hline
\multirow{5}{*}{delicious} 
&	$20\%$ & {\bf 61.20}& 53.68& 57.27& 53.01& 61.13& 42.87\\
&	$40\%$ & {\bf 61.23}& 49.13& 52.95& 56.20& 61.08& 42.05\\
&	$60\%$ & {\bf 61.15}& 46.76& 49.58& 57.07& 61.09& 42.22\\
&	$80\%$ & {\bf 61.13}& 48.06& 47.34& 57.09& 61.09& 42.01\\
&	$100\%$ & {\bf 61.12}& 46.11& 45.92& 57.91& 61.09& 41.34\\
\end{tabular}
\end{table*}
\begin{table*}[h]
\caption{Comparison for dimensionality reductions approach on fully observed $Y$ with various rank. SQ for squared loss, LR for logistic loss, SH for squared hinge loss, and WAR for weighted approximated-rank loss}
\label{tab:Y-o-1-p-5}
\centering
\small
\begin{tabular}{lr|r|r|r|r|r|r}
	&	& \multicolumn{6}{@{}c@{}}{Top-$5$ Accuracy}\\
	&	& \multicolumn{3}{@{}c@{}|}{\MLR} & \BCS & \CPLST & \WSABIE \\
	&$k/L$ & SQ & LR & SH & SQ& SQ & WAR \\
\hline
\hline
\multirow{5}{*}{bibtex} 
&	$20\%$ & {\bf 24.49}& 19.24& 20.33& 15.39& 23.11& 21.92\\
&	$40\%$ & {\bf 26.84}& 20.61& 18.54& 19.95& 24.96& 22.47\\
&	$60\%$ & {\bf 27.66}& 20.99& 17.61& 22.43& 26.07& 23.33\\
&	$80\%$ & {\bf 28.20}& 21.48& 17.46& 24.07& 26.47& 23.44\\
&	$100\%$ & {\bf 28.01}& 22.03& 16.83& 24.48& 26.47& 23.44\\
\hline
\multirow{5}{*}{autofood} 
&	$20\%$ & {\bf 81.05}& 80.00& 75.79& 44.21& 36.84& 66.32\\
&	$40\%$ & 73.68& {\bf 78.42}& 76.84& 51.05& 36.32& 66.84\\
&	$60\%$ & 69.47& {\bf 78.95}& 78.42& 57.37& 36.32& 60.53\\
&	$80\%$ & 74.74& 78.95& {\bf 80.53}& 68.95& 36.84& 66.84\\
&	$100\%$ & 72.63& 78.42& {\bf 83.16}& 62.11& 36.84& 61.58\\
\hline
\multirow{5}{*}{compphys} 
&	$20\%$ & 72.00& {\bf 73.50}& 72.50& 32.50& 37.50& 46.00\\
&	$40\%$ & 73.00& 74.00& {\bf 74.50}& 54.50& 35.50& 41.00\\
&	$60\%$ & 73.00& {\bf 74.00}& {\bf 74.00}& 43.50& 34.50& 44.00\\
&	$80\%$ & 73.00& 73.00& {\bf 74.00}& 47.50& 36.00& 46.50\\
&	$100\%$ & 72.50& 72.50& {\bf 73.00}& 54.50& 36.00& 49.50\\
\hline
\multirow{5}{*}{delicious} 
&	$20\%$ & {\bf 56.46}& 49.46& 52.94& 47.91& 56.30& 39.79\\
&	$40\%$ & {\bf 56.39}& 45.66& 49.54& 51.61& 56.28& 39.27\\
&	$60\%$ & {\bf 56.28}& 43.22& 46.93& 52.85& 56.23& 38.97\\
&	$80\%$ & {\bf 56.27}& 44.03& 45.43& 52.92& 56.23& 39.27\\
&	$100\%$ & {\bf 56.27}& 42.11& 44.24& 53.28& 56.23& 38.41\\
\end{tabular}
\end{table*}
\begin{table*}[h]
\caption{Comparison for dimensionality reductions approach on fully observed $Y$ with various rank. SQ for squared loss, LR for logistic loss, SH for squared hinge loss, and WAR for weighted approximated-rank loss}
\label{tab:Y-o-1-p-0}
\centering
\small
\begin{tabular}{lr|r|r|r|r|r}
	&	& \multicolumn{5}{@{}c@{}}{Hamming Loss}\\
	&	& \multicolumn{3}{@{}c@{}|}{\MLR} & \BCS & \CPLST \\
	&$k/L$ & SQ & LR & SH & SQ& SQ \\
\hline
\hline
\multirow{5}{*}{bibtex} 
&	$20\%$ & {\bf 0.0126}& 0.0211& 0.0231& 0.0150& {\bf 0.0127}\\
&	$40\%$ & {\bf 0.0124}& 0.0240& 0.0285& 0.0140& 0.0126\\
&	$60\%$ & {\bf 0.0123}& 0.0233& 0.0320& 0.0132& 0.0126\\
&	$80\%$ & {\bf 0.0123}& 0.0242& 0.0343& 0.0130& 0.0125\\
&	$100\%$ & {\bf 0.0122}& 0.0236& 0.0375& 0.0129& 0.0125\\
\hline
\multirow{5}{*}{autofood} 
&	$20\%$ & {\bf 0.0547}& 0.0621& 0.0588& 0.0846& 0.0996\\
&	$40\%$ & 0.0590& 0.0608& {\bf 0.0578}& 0.0846& 0.0975\\
&	$60\%$ & 0.0593& 0.0611& {\bf 0.0586}& 0.0838& 0.0945\\
&	$80\%$ & 0.0572& 0.0611& {\bf 0.0569}& 1.0000& 0.0944\\
&	$100\%$ & 0.0603& 0.0617& {\bf 0.0586}& 1.0000& 0.0944\\
\hline
\multirow{5}{*}{compphys} 
&	$20\%$ & 0.0457& 0.0470& {\bf 0.0456}& 0.0569& 0.0530\\
&	$40\%$ & {\bf 0.0454}& 0.0466& 0.0456& 0.0569& 0.0526\\
&	$60\%$ & {\bf 0.0454}& 0.0469& 0.0460& 0.0569& 0.0530\\
&	$80\%$ & 0.0464& 0.0484& {\bf 0.0456}& 0.0569& 0.0755\\
&	$100\%$ & 0.0453& 0.0469& {\bf 0.0450}& 0.0569& 0.0755\\
\hline
\multirow{5}{*}{delicious} 
&	$20\%$ & {\bf 0.0181}& 0.0196& 0.0187& 0.0189& {\bf 0.0182}\\
&	$40\%$ & {\bf 0.0181}& 0.0221& 0.0198& 0.0186& {\bf 0.0182}\\
&	$60\%$ & {\bf 0.0182}& 0.0239& 0.0207& 0.0187& {\bf 0.0182}\\
&	$80\%$ & {\bf 0.0182}& 0.0253& 0.0212& 0.0186& {\bf 0.0182}\\
&	$100\%$ & {\bf 0.0182}& 0.0260& 0.0216& 0.0186& {\bf 0.0182}\\
\end{tabular}
\end{table*}
\begin{table*}[h]
\caption{Comparison for dimensionality reductions approach on fully observed $Y$ with various rank. SQ for squared loss, LR for logistic loss, SH for squared hinge loss, and WAR for weighted approximated-rank loss}
\label{tab:Y-o-1-p--1}
\centering
\small
\begin{tabular}{lr|r|r|r|r|r|r}
	&	& \multicolumn{6}{@{}c@{}}{Average AUC}\\
	&	& \multicolumn{3}{@{}c@{}|}{\MLR} & \BCS & \CPLST & \WSABIE \\
	&$k/L$ & SQ & LR & SH & SQ& SQ & WAR \\
\hline
\hline
\multirow{5}{*}{bibtex} 
&	$20\%$ & 0.8910& 0.8677& 0.8541& 0.7875& 0.8657& {\bf 0.9055}\\
&	$40\%$ & 0.9015& 0.8809& 0.8467& 0.8263& 0.8802& {\bf 0.9092}\\
&	$60\%$ & 0.9040& 0.8861& 0.8505& 0.8468& 0.8854& {\bf 0.9089}\\
&	$80\%$ & 0.9035& 0.8875& 0.8491& 0.8560& 0.8882& {\bf 0.9164}\\
&	$100\%$ & 0.9024& 0.8915& 0.8419& 0.8614& 0.8878& {\bf 0.9182}\\
\hline
\multirow{5}{*}{autofood} 
&	$20\%$ & 0.9565& {\bf 0.9598}& 0.9424& 0.7599& 0.7599& 0.8779\\
&	$40\%$ & 0.9277& {\bf 0.9590}& 0.9485& 0.7994& 0.7501& 0.8806\\
&	$60\%$ & 0.8815& {\bf 0.9582}& 0.9513& 0.8282& 0.7552& 0.8518\\
&	$80\%$ & 0.9280& {\bf 0.9588}& 0.9573& 0.8611& 0.7538& 0.8520\\
&	$100\%$ & 0.9361& {\bf 0.9581}& 0.9561& 0.8718& 0.7539& 0.8471\\
\hline
\multirow{5}{*}{compphys} 
&	$20\%$ & 0.9163& 0.9223& {\bf 0.9274}& 0.6972& 0.7692& 0.8212\\
&	$40\%$ & {\bf 0.9199}& 0.9157& 0.9191& 0.7881& 0.7742& 0.8066\\
&	$60\%$ & {\bf 0.9179}& 0.9143& 0.9098& 0.7705& 0.7705& 0.8040\\
&	$80\%$ & 0.9187& 0.9003& {\bf 0.9220}& 0.7820& 0.7806& 0.7742\\
&	$100\%$ & {\bf 0.9205}& 0.9040& 0.8977& 0.7884& 0.7804& 0.7951\\
\hline
\multirow{5}{*}{delicious} 
&	$20\%$ & 0.8854& 0.8588& {\bf 0.8894}& 0.7308& 0.8833& 0.8561\\
&	$40\%$ & 0.8827& 0.8534& {\bf 0.8868}& 0.7635& 0.8814& 0.8553\\
&	$60\%$ & 0.8814& 0.8517& {\bf 0.8852}& 0.7842& 0.8834& 0.8523\\
&	$80\%$ & 0.8814& 0.8468& {\bf 0.8845}& 0.7941& 0.8834& 0.8558\\
&	$100\%$ & 0.8814& 0.8404& {\bf 0.8836}& 0.8000& 0.8834& 0.8557\\
\end{tabular}
\end{table*}

\subsection{Detailed Results with Missing Labels}
\label{app:exp-missing}
\begin{itemize}
  \item Table \ref{tab:Y-missing-p-1} shows the top-1 accuracy results for the 
    case with various missing ratios and dimension reduction rates. 
  \item Table \ref{tab:Y-missing-p-3} shows the top-3 accuracy results for the 
    case with various missing ratios and dimension reduction rates. 
  \item Table \ref{tab:Y-missing-p-5} shows the top-5 accuracy results for the 
    case with various missing ratios and dimension reduction rates. 
  \item Table \ref{tab:Y-missing-p-0} shows the Hamming loss results for the 
    case with various missing ratios and dimension reduction rates. 
  \item Table \ref{tab:Y-missing-p--1} shows the average AUC results for the 
    case with various missing ratios and dimension reduction rates. 
\end{itemize}
\begin{table}[h]
\caption{Comparison for $Y$ with missing labels}
\label{tab:Y-missing-p-1}
\centering
\small
{
\begin{tabular}{lrr|rrr|rr|rr}
	& & &\multicolumn{7}{@{}c@{}}{Top-$1$ Accuracy} \\
\multirow{2}{*}{dataset}& \multirow{2}{*}{$\frac{k}{L}$}& \multirow{2}{*}{$\frac{|\Omega|}{nL}$}
 &\multicolumn{3}{c|}{Squared}& \multicolumn{2}{c|}{Logsitic}& \multicolumn{2}{c}{Squared Hinge}\\
	&	&	& \MLR & \BCS & \BR & \MLR & \BR & \MLR & \BR\\
\hline \hline
\multirow{12}{*}{bibtex}&\multirow{4}{*}{$20\%$}
& $5\%$& 30.30& 30.22& {\bf 42.90}& 41.51& {\bf 46.68}& 30.42& {\bf 44.97}\\
&& $10\%$& 39.84& 33.56& {\bf 44.53}& 41.99& {\bf 51.09}& 33.44& {\bf 48.55}\\
&& $20\%$& {\bf 48.35}& 40.12& 46.08& 43.06& {\bf 55.94}& 37.22& {\bf 52.84}\\
&& $40\%$& {\bf 52.37}& 41.79& 43.82& 42.27& {\bf 58.57}& 40.24& {\bf 55.39}\\
\cline{2-10}
&\multirow{4}{*}{$40\%$}
& $5\%$& 34.35& 39.17& {\bf 42.90}& 43.42& {\bf 46.68}& 31.13& {\bf 44.97}\\
&& $10\%$& 42.11& 39.96& {\bf 44.53}& 46.00& {\bf 51.09}& 29.03& {\bf 48.55}\\
&& $20\%$& {\bf 51.97}& 45.49& 46.08& 47.40& {\bf 55.94}& 32.05& {\bf 52.84}\\
&& $40\%$& {\bf 56.38}& 50.10& 43.82& 49.70& {\bf 58.57}& 38.17& {\bf 55.39}\\
\cline{2-10}
&\multirow{4}{*}{$60\%$}
& $5\%$& 36.58& 41.87& {\bf 42.90}& 43.54& {\bf 46.68}& 42.54& {\bf 44.97}\\
&& $10\%$& {\bf 45.53}& 45.13& 44.53& 39.36& {\bf 51.09}& 31.37& {\bf 48.55}\\
&& $20\%$& {\bf 53.52}& 49.54& 46.08& 46.12& {\bf 55.94}& 33.28& {\bf 52.84}\\
&& $40\%$& {\bf 57.18}& 54.19& 43.82& 48.83& {\bf 58.57}& 32.13& {\bf 55.39}\\
\hline \hline
\multirow{12}{*}{autofood}&\multirow{4}{*}{$20\%$}
& $5\%$& {\bf 7.89}& 0.00& {\bf 7.89}& {\bf 7.89}& {\bf 7.89}& {\bf 7.89}& {\bf 7.89}\\
&& $10\%$& 44.74& 2.63& {\bf 50.00}& {\bf 55.26}& 44.74& {\bf 50.00}& {\bf 50.00}\\
&& $20\%$& {\bf 63.16}& 0.00& 57.89& {\bf 73.68}& 47.37& {\bf 68.42}& 57.89\\
&& $40\%$& 60.53& 15.79& {\bf 78.95}& {\bf 81.58}& 68.42& {\bf 86.84}& 78.95\\
\cline{2-10}
&\multirow{4}{*}{$40\%$}
& $5\%$& {\bf 10.53}& {\bf 10.53}& 7.89& {\bf 7.89}& {\bf 7.89}& {\bf 13.16}& 7.89\\
&& $10\%$& {\bf 57.89}& 7.89& 50.00& {\bf 60.53}& 44.74& {\bf 55.26}& 50.00\\
&& $20\%$& {\bf 76.32}& 31.58& 57.89& {\bf 78.95}& 47.37& {\bf 76.32}& 57.89\\
&& $40\%$& 60.53& 5.26& {\bf 78.95}& {\bf 84.21}& 68.42& {\bf 84.21}& 78.95\\
\cline{2-10}
&\multirow{4}{*}{$60\%$}
& $5\%$& 7.89& {\bf 10.53}& 7.89& {\bf 7.89}& {\bf 7.89}& {\bf 7.89}& {\bf 7.89}\\
&& $10\%$& {\bf 57.89}& 23.68& 50.00& {\bf 57.89}& 44.74& {\bf 55.26}& 50.00\\
&& $20\%$& {\bf 73.68}& 57.89& 57.89& {\bf 78.95}& 47.37& {\bf 76.32}& 57.89\\
&& $40\%$& 63.16& 36.84& {\bf 78.95}& {\bf 81.58}& 68.42& {\bf 89.47}& 78.95\\
\hline \hline
\multirow{12}{*}{compphys}&\multirow{4}{*}{$20\%$}
& $5\%$& {\bf 62.50}& 35.00& 42.50& {\bf 45.00}& {\bf 45.00}& {\bf 67.50}& 42.50\\
&& $10\%$& {\bf 75.00}& 10.00& 52.50& {\bf 67.50}& 52.50& {\bf 55.00}& 52.50\\
&& $20\%$& {\bf 72.50}& 7.50& 52.50& {\bf 72.50}& 52.50& {\bf 70.00}& 52.50\\
&& $40\%$& {\bf 87.50}& 5.00& 52.50& {\bf 77.50}& 52.50& {\bf 80.00}& 52.50\\
\cline{2-10}
&\multirow{4}{*}{$40\%$}
& $5\%$& {\bf 65.00}& 60.00& 42.50& {\bf 45.00}& {\bf 45.00}& {\bf 65.00}& 42.50\\
&& $10\%$& {\bf 70.00}& 17.50& 52.50& {\bf 65.00}& 52.50& {\bf 72.50}& 52.50\\
&& $20\%$& {\bf 72.50}& 52.50& 52.50& {\bf 70.00}& 52.50& {\bf 75.00}& 52.50\\
&& $40\%$& {\bf 80.00}& 42.50& 52.50& {\bf 80.00}& 52.50& {\bf 80.00}& 52.50\\
\cline{2-10}
&\multirow{4}{*}{$60\%$}
& $5\%$& {\bf 67.50}& 52.50& 42.50& {\bf 45.00}& {\bf 45.00}& {\bf 65.00}& 42.50\\
&& $10\%$& {\bf 70.00}& 52.50& 52.50& {\bf 67.50}& 52.50& {\bf 67.50}& 52.50\\
&& $20\%$& {\bf 77.50}& 52.50& 52.50& {\bf 80.00}& 52.50& {\bf 80.00}& 52.50\\
&& $40\%$& {\bf 82.50}& 52.50& 52.50& {\bf 80.00}& 52.50& {\bf 80.00}& 52.50\\
\end{tabular}
}
\end{table}
\begin{table}[h]
\caption{Comparison for $Y$ with missing labels}
\label{tab:Y-missing-p-3}
\centering
\small

{
\begin{tabular}{lrr|rrr|rr|rr}
	& & &\multicolumn{7}{@{}c@{}}{Top-$3$ Accuracy} \\
\multirow{2}{*}{dataset}& \multirow{2}{*}{$\frac{k}{L}$}& \multirow{2}{*}{$\frac{|\Omega|}{nL}$}
 &\multicolumn{3}{c|}{Squared}& \multicolumn{2}{c|}{Logsitic}& \multicolumn{2}{c}{Squared Hinge}\\
	&	&	& \MLR & \BCS & \BR & \MLR & \BR & \MLR & \BR\\
\hline \hline
\multirow{12}{*}{bibtex}&\multirow{4}{*}{$20\%$}
& $5\%$& 16.06& 14.29& {\bf 22.19}& 21.74& {\bf 24.47}& 16.10& {\bf 23.29}\\
&& $10\%$& 20.95& 16.29& {\bf 24.10}& 22.88& {\bf 28.43}& 17.64& {\bf 26.69}\\
&& $20\%$& {\bf 26.34}& 18.78& 25.78& 23.21& {\bf 31.92}& 21.06& {\bf 29.56}\\
&& $40\%$& {\bf 30.17}& 21.55& 26.26& 23.61& {\bf 34.50}& 23.05& {\bf 31.99}\\
\cline{2-10}
&\multirow{4}{*}{$40\%$}
& $5\%$& 18.73& 18.99& {\bf 22.19}& 22.84& {\bf 24.47}& 17.03& {\bf 23.29}\\
&& $10\%$& 22.49& 20.16& {\bf 24.10}& 25.18& {\bf 28.43}& 16.62& {\bf 26.69}\\
&& $20\%$& {\bf 28.50}& 23.84& 25.78& 25.79& {\bf 31.92}& 18.97& {\bf 29.56}\\
&& $40\%$& {\bf 32.74}& 27.58& 26.26& 27.18& {\bf 34.50}& 21.18& {\bf 31.99}\\
\cline{2-10}
&\multirow{4}{*}{$60\%$}
& $5\%$& 18.81& 21.09& {\bf 22.19}& 22.62& {\bf 24.47}& 22.48& {\bf 23.29}\\
&& $10\%$& 23.96& 24.06& {\bf 24.10}& 19.84& {\bf 28.43}& 17.28& {\bf 26.69}\\
&& $20\%$& {\bf 29.07}& 27.05& 25.78& 25.13& {\bf 31.92}& 19.14& {\bf 29.56}\\
&& $40\%$& {\bf 33.55}& 31.13& 26.26& 27.66& {\bf 34.50}& 19.46& {\bf 31.99}\\
\hline \hline
\multirow{12}{*}{autofood}&\multirow{4}{*}{$20\%$}
& $5\%$& {\bf 30.70}& 11.40& 19.30& {\bf 29.82}& 17.54& {\bf 38.60}& 19.30\\
&& $10\%$& {\bf 52.63}& 5.26& 33.33& {\bf 50.88}& 23.68& {\bf 57.02}& 33.33\\
&& $20\%$& 59.65& 10.53& {\bf 62.28}& {\bf 70.18}& 53.51& {\bf 66.67}& 61.40\\
&& $40\%$& 57.89& 20.18& {\bf 71.93}& {\bf 76.32}& 63.16& {\bf 75.44}& 71.93\\
\cline{2-10}
&\multirow{4}{*}{$40\%$}
& $5\%$& {\bf 26.32}& 15.79& 19.30& {\bf 29.82}& 17.54& {\bf 31.58}& 19.30\\
&& $10\%$& {\bf 59.65}& 12.28& 33.33& {\bf 51.75}& 23.68& {\bf 53.51}& 33.33\\
&& $20\%$& {\bf 67.54}& 35.09& 62.28& {\bf 71.05}& 53.51& {\bf 64.04}& 61.40\\
&& $40\%$& 55.26& 33.33& {\bf 71.93}& {\bf 78.07}& 63.16& {\bf 77.19}& 71.93\\
\cline{2-10}
&\multirow{4}{*}{$60\%$}
& $5\%$& {\bf 25.44}& 8.77& 19.30& {\bf 28.95}& 17.54& {\bf 22.81}& 19.30\\
&& $10\%$& {\bf 52.63}& 35.09& 33.33& {\bf 50.00}& 23.68& {\bf 61.40}& 33.33\\
&& $20\%$& {\bf 68.42}& 35.09& 62.28& {\bf 73.68}& 53.51& {\bf 71.05}& 61.40\\
&& $40\%$& 57.02& 23.68& {\bf 71.93}& {\bf 75.44}& 63.16& {\bf 74.56}& 71.93\\
\hline \hline
\multirow{12}{*}{compphys}&\multirow{4}{*}{$20\%$}
& $5\%$& {\bf 46.67}& 32.50& 28.33& {\bf 40.00}& 28.33& {\bf 40.00}& 28.33\\
&& $10\%$& {\bf 53.33}& 9.17& 37.50& {\bf 59.17}& 29.17& {\bf 40.83}& 37.50\\
&& $20\%$& {\bf 62.50}& 10.83& 31.67& {\bf 60.83}& 28.33& {\bf 61.67}& 31.67\\
&& $40\%$& {\bf 69.17}& 26.67& 43.33& {\bf 73.33}& 33.33& {\bf 70.83}& 43.33\\
\cline{2-10}
&\multirow{4}{*}{$40\%$}
& $5\%$& {\bf 45.83}& 27.50& 28.33& {\bf 37.50}& 28.33& {\bf 41.67}& 28.33\\
&& $10\%$& {\bf 57.50}& 20.83& 37.50& {\bf 60.00}& 29.17& {\bf 55.83}& 37.50\\
&& $20\%$& {\bf 65.00}& 35.83& 31.67& {\bf 60.00}& 28.33& {\bf 61.67}& 31.67\\
&& $40\%$& {\bf 68.33}& 32.50& 43.33& {\bf 70.83}& 33.33& {\bf 73.33}& 43.33\\
\cline{2-10}
&\multirow{4}{*}{$60\%$}
& $5\%$& {\bf 45.00}& 30.83& 28.33& {\bf 35.83}& 28.33& {\bf 45.00}& 28.33\\
&& $10\%$& {\bf 59.17}& 26.67& 37.50& {\bf 61.67}& 29.17& {\bf 56.67}& 37.50\\
&& $20\%$& {\bf 65.00}& 29.17& 31.67& {\bf 60.83}& 28.33& {\bf 64.17}& 31.67\\
&& $40\%$& {\bf 71.67}& 30.00& 43.33& {\bf 65.83}& 33.33& {\bf 70.83}& 43.33\\
\end{tabular}
}
\end{table}
\begin{table}[h]
\caption{Comparison for $Y$ with missing labels}
\label{tab:Y-missing-p-5}
\centering
\small
{
\begin{tabular}{lrr|rrr|rr|rr}
	& & &\multicolumn{7}{@{}c@{}}{Top-$5$ Accuracy} \\
\multirow{2}{*}{dataset}& \multirow{2}{*}{$\frac{k}{L}$}& \multirow{2}{*}{$\frac{|\Omega|}{nL}$}
 &\multicolumn{3}{c|}{Squared}& \multicolumn{2}{c|}{Logsitic}& \multicolumn{2}{c}{Squared Hinge}\\
	&	&	& \MLR & \BCS & \BR & \MLR & \BR & \MLR & \BR\\
\hline \hline
\multirow{12}{*}{bibtex}&\multirow{4}{*}{$20\%$}
& $5\%$& 11.71& 10.32& {\bf 16.14}& 16.34& {\bf 17.74}& 12.07& {\bf 17.32}\\
&& $10\%$& 15.42& 11.55& {\bf 17.77}& 16.91& {\bf 20.80}& 13.11& {\bf 19.65}\\
&& $20\%$& {\bf 19.51}& 13.26& 18.81& 17.07& {\bf 23.95}& 15.52& {\bf 22.12}\\
&& $40\%$& {\bf 22.05}& 15.32& 19.13& 17.55& {\bf 25.57}& 17.57& {\bf 23.30}\\
\cline{2-10}
&\multirow{4}{*}{$40\%$}
& $5\%$& 13.53& 13.25& {\bf 16.14}& 17.02& {\bf 17.74}& 12.70& {\bf 17.32}\\
&& $10\%$& 16.25& 14.30& {\bf 17.77}& 18.78& {\bf 20.80}& 12.24& {\bf 19.65}\\
&& $20\%$& {\bf 20.56}& 17.36& 18.81& 19.05& {\bf 23.95}& 14.46& {\bf 22.12}\\
&& $40\%$& {\bf 23.75}& 19.73& 19.13& 19.67& {\bf 25.57}& 15.86& {\bf 23.30}\\
\cline{2-10}
&\multirow{4}{*}{$60\%$}
& $5\%$& 13.61& 14.78& {\bf 16.14}& 16.62& {\bf 17.74}& 16.56& {\bf 17.32}\\
&& $10\%$& 16.99& 17.31& {\bf 17.77}& 14.41& {\bf 20.80}& 12.91& {\bf 19.65}\\
&& $20\%$& {\bf 21.10}& 19.51& 18.81& 18.23& {\bf 23.95}& 14.17& {\bf 22.12}\\
&& $40\%$& {\bf 24.50}& 22.31& 19.13& 20.38& {\bf 25.57}& 14.95& {\bf 23.30}\\
\hline \hline
\multirow{12}{*}{autofood}&\multirow{4}{*}{$20\%$}
& $5\%$& {\bf 35.26}& 8.42& 25.26& {\bf 34.21}& 21.58& {\bf 36.84}& 25.26\\
&& $10\%$& {\bf 46.84}& 6.84& 35.79& {\bf 51.05}& 32.11& {\bf 48.95}& 35.79\\
&& $20\%$& 50.53& 10.53& {\bf 57.89}& {\bf 66.84}& 52.11& {\bf 60.53}& 57.89\\
&& $40\%$& 52.11& 16.84& {\bf 68.42}& {\bf 73.16}& 56.32& {\bf 72.11}& 68.42\\
\cline{2-10}
&\multirow{4}{*}{$40\%$}
& $5\%$& {\bf 32.11}& 17.89& 25.26& {\bf 31.58}& 21.58& {\bf 30.00}& 25.26\\
&& $10\%$& {\bf 49.47}& 10.00& 35.79& {\bf 50.53}& 32.11& {\bf 45.26}& 35.79\\
&& $20\%$& {\bf 64.74}& 32.11& 57.89& {\bf 66.32}& 52.11& {\bf 60.53}& 57.89\\
&& $40\%$& 50.53& 28.95& {\bf 68.42}& {\bf 73.16}& 56.32& {\bf 74.74}& 68.42\\
\cline{2-10}
&\multirow{4}{*}{$60\%$}
& $5\%$& {\bf 31.58}& 17.37& 25.26& {\bf 31.05}& 21.58& {\bf 30.00}& 25.26\\
&& $10\%$& {\bf 50.53}& 31.58& 35.79& {\bf 52.63}& 32.11& {\bf 53.68}& 35.79\\
&& $20\%$& {\bf 64.74}& 28.95& 57.89& {\bf 68.42}& 52.11& {\bf 67.89}& 57.89\\
&& $40\%$& 58.95& 20.00& {\bf 68.42}& {\bf 71.58}& 56.32& {\bf 69.47}& 68.42\\
\hline \hline
\multirow{12}{*}{compphys}&\multirow{4}{*}{$20\%$}
& $5\%$& {\bf 34.50}& 23.00& 25.00& {\bf 28.50}& 26.00& {\bf 34.50}& 25.00\\
&& $10\%$& {\bf 50.50}& 13.50& 28.50& {\bf 51.50}& 24.00& {\bf 41.50}& 29.00\\
&& $20\%$& {\bf 52.00}& 11.50& 36.50& {\bf 55.00}& 30.00& {\bf 53.00}& 36.50\\
&& $40\%$& {\bf 60.50}& 24.00& 38.00& {\bf 64.50}& 31.00& {\bf 64.00}& 38.50\\
\cline{2-10}
&\multirow{4}{*}{$40\%$}
& $5\%$& {\bf 34.50}& 22.00& 25.00& {\bf 29.50}& 26.00& {\bf 33.50}& 25.00\\
&& $10\%$& {\bf 53.50}& 29.00& 28.50& {\bf 51.50}& 24.00& {\bf 46.00}& 29.00\\
&& $20\%$& {\bf 56.50}& 31.00& 36.50& {\bf 55.50}& 30.00& {\bf 52.50}& 36.50\\
&& $40\%$& {\bf 59.50}& 26.00& 38.00& {\bf 61.50}& 31.00& {\bf 62.50}& 38.50\\
\cline{2-10}
&\multirow{4}{*}{$60\%$}
& $5\%$& {\bf 36.00}& 22.00& 25.00& {\bf 27.50}& 26.00& {\bf 33.50}& 25.00\\
&& $10\%$& {\bf 53.00}& 24.50& 28.50& {\bf 50.50}& 24.00& {\bf 50.50}& 29.00\\
&& $20\%$& {\bf 56.50}& 29.00& 36.50& {\bf 54.00}& 30.00& {\bf 55.50}& 36.50\\
&& $40\%$& {\bf 61.00}& 32.00& 38.00& {\bf 61.50}& 31.00& {\bf 63.50}& 38.50\\
\end{tabular}
}
\end{table}
\begin{table}[h]
\caption{Comparison for $Y$ with missing labels}
\label{tab:Y-missing-p-0}
\centering
\small
{
\begin{tabular}{lrr|rrr|rr|rr}
	& & &\multicolumn{7}{@{}c@{}}{Hamming Loss} \\
\multirow{2}{*}{dataset}& \multirow{2}{*}{$\frac{k}{L}$}& \multirow{2}{*}{$\frac{|\Omega|}{nL}$}
 &\multicolumn{3}{c|}{Squared}& \multicolumn{2}{c|}{Logsitic}& \multicolumn{2}{c}{Squared Hinge}\\
	&	&	& \MLR & \BCS & \BR & \MLR & \BR & \MLR & \BR\\
\hline \hline
\multirow{12}{*}{bibtex}&\multirow{4}{*}{$20\%$}
& $5\%$& 0.0158& 0.1480& {\bf 0.0144}& 0.0143& {\bf 0.0138}& 0.0180& {\bf 0.0137}\\
&& $10\%$& {\bf 0.0146}& 0.1360& 0.0156& 0.0144& {\bf 0.0134}& 0.0187& {\bf 0.0135}\\
&& $20\%$& {\bf 0.0136}& 0.1179& 0.0193& 0.0156& {\bf 0.0132}& 0.0210& {\bf 0.0136}\\
&& $40\%$& {\bf 0.0131}& 0.0994& 0.0251& 0.0174& {\bf 0.0128}& 0.0242& {\bf 0.0141}\\
\cline{2-10}
&\multirow{4}{*}{$40\%$}
& $5\%$& 0.0152& 0.2837& {\bf 0.0144}& 0.0141& {\bf 0.0138}& 0.0175& {\bf 0.0137}\\
&& $10\%$& {\bf 0.0149}& 0.2716& 0.0156& 0.0141& {\bf 0.0134}& 0.0211& {\bf 0.0135}\\
&& $20\%$& {\bf 0.0136}& 0.2496& 0.0193& 0.0150& {\bf 0.0132}& 0.0226& {\bf 0.0136}\\
&& $40\%$& {\bf 0.0128}& 0.2271& 0.0251& 0.0160& {\bf 0.0128}& 0.0269& {\bf 0.0141}\\
\cline{2-10}
&\multirow{4}{*}{$60\%$}
& $5\%$& 0.0154& 0.4082& {\bf 0.0144}& 0.0145& {\bf 0.0138}& 0.0154& {\bf 0.0137}\\
&& $10\%$& {\bf 0.0147}& 0.3978& 0.0156& 0.0163& {\bf 0.0134}& 0.0215& {\bf 0.0135}\\
&& $20\%$& {\bf 0.0138}& 0.3726& 0.0193& 0.0157& {\bf 0.0132}& 0.0252& {\bf 0.0136}\\
&& $40\%$& {\bf 0.0129}& 0.3638& 0.0251& 0.0172& {\bf 0.0128}& 0.0312& {\bf 0.0141}\\
\hline \hline
\multirow{12}{*}{autofood}&\multirow{4}{*}{$20\%$}
& $5\%$& {\bf 0.0924}& 0.1727& 0.0942& {\bf 0.0918}& 0.0991& {\bf 0.0884}& 0.0942\\
&& $10\%$& {\bf 0.0807}& 0.1449& 0.0837& {\bf 0.0832}& 0.0854& {\bf 0.0811}& 0.0837\\
&& $20\%$& {\bf 0.0750}& 0.1436& 0.0760& {\bf 0.0686}& 0.0843& {\bf 0.0697}& 0.0760\\
&& $40\%$& 0.0780& 0.1399& {\bf 0.0752}& {\bf 0.0655}& 0.0838& {\bf 0.0629}& 0.0750\\
\cline{2-10}
&\multirow{4}{*}{$40\%$}
& $5\%$& {\bf 0.0919}& 0.2887& 0.0942& {\bf 0.0919}& 0.0991& {\bf 0.0941}& 0.0942\\
&& $10\%$& {\bf 0.0801}& 0.2264& 0.0837& {\bf 0.0812}& 0.0854& {\bf 0.0814}& 0.0837\\
&& $20\%$& {\bf 0.0671}& 0.2445& 0.0760& {\bf 0.0681}& 0.0843& {\bf 0.0697}& 0.0760\\
&& $40\%$& 0.0903& 0.2042& {\bf 0.0752}& {\bf 0.0647}& 0.0838& {\bf 0.0648}& 0.0750\\
\cline{2-10}
&\multirow{4}{*}{$60\%$}
& $5\%$& {\bf 0.0932}& 0.4189& 0.0942& {\bf 0.0921}& 0.0991& {\bf 0.0937}& 0.0942\\
&& $10\%$& 0.0840& 0.4144& {\bf 0.0837}& {\bf 0.0817}& 0.0854& {\bf 0.0817}& 0.0837\\
&& $20\%$& {\bf 0.0689}& 0.3596& 0.0760& {\bf 0.0676}& 0.0843& {\bf 0.0692}& 0.0760\\
&& $40\%$& {\bf 0.0724}& 0.3384& 0.0752& {\bf 0.0650}& 0.0838& {\bf 0.0645}& 0.0750\\
\hline \hline
\multirow{12}{*}{compphys}&\multirow{4}{*}{$20\%$}
& $5\%$& {\bf 0.0555}& 0.1391& 0.0556& {\bf 0.0554}& 0.0555& 0.0567& {\bf 0.0556}\\
&& $10\%$& {\bf 0.0536}& 0.1446& 0.0565& {\bf 0.0542}& 0.0569& {\bf 0.0554}& 0.0565\\
&& $20\%$& {\bf 0.0524}& 0.1431& 0.0566& {\bf 0.0518}& 0.0566& {\bf 0.0518}& 0.0566\\
&& $40\%$& {\bf 0.0484}& 0.1048& 0.0543& {\bf 0.0489}& 0.0561& {\bf 0.0488}& 0.0543\\
\cline{2-10}
&\multirow{4}{*}{$40\%$}
& $5\%$& 0.0567& 0.2924& {\bf 0.0556}& {\bf 0.0555}& {\bf 0.0555}& 0.0566& {\bf 0.0556}\\
&& $10\%$& {\bf 0.0532}& 0.2532& 0.0565& {\bf 0.0535}& 0.0569& {\bf 0.0532}& 0.0565\\
&& $20\%$& {\bf 0.0518}& 0.2569& 0.0566& {\bf 0.0513}& 0.0566& {\bf 0.0518}& 0.0566\\
&& $40\%$& {\bf 0.0505}& 0.1766& 0.0543& {\bf 0.0495}& 0.0561& {\bf 0.0484}& 0.0543\\
\cline{2-10}
&\multirow{4}{*}{$60\%$}
& $5\%$& 0.0558& 0.4394& {\bf 0.0556}& 0.0556& {\bf 0.0555}& {\bf 0.0555}& 0.0556\\
&& $10\%$& {\bf 0.0532}& 0.4148& 0.0565& {\bf 0.0532}& 0.0569& {\bf 0.0544}& 0.0565\\
&& $20\%$& {\bf 0.0516}& 0.3797& 0.0566& {\bf 0.0519}& 0.0566& {\bf 0.0517}& 0.0566\\
&& $40\%$& {\bf 0.0486}& 0.3563& 0.0543& {\bf 0.0495}& 0.0561& {\bf 0.0480}& 0.0543\\
\end{tabular}
}
\end{table}
\begin{table}[h]
\caption{Comparison for $Y$ with missing labels}
\label{tab:Y-missing-p--1}
\centering
\small
{
\begin{tabular}{lrr|rrr|rr|rr}
	& & &\multicolumn{7}{@{}c@{}}{Average AUC} \\
\multirow{2}{*}{dataset}& \multirow{2}{*}{$\frac{k}{L}$}& \multirow{2}{*}{$\frac{|\Omega|}{nL}$}
 &\multicolumn{3}{c|}{Squared}& \multicolumn{2}{c|}{Logsitic}& \multicolumn{2}{c}{Squared Hinge}\\
	&	&	& \MLR & \BCS & \BR & \MLR & \BR & \MLR & \BR\\
\hline \hline
\multirow{12}{*}{bibtex}&\multirow{4}{*}{$20\%$}
& $5\%$& 0.7115& 0.6529& {\bf 0.7789}& 0.8066& {\bf 0.8123}& 0.7363& {\bf 0.7998}\\
&& $10\%$& 0.7665& 0.6756& {\bf 0.7954}& 0.8208& {\bf 0.8561}& 0.7371& {\bf 0.8210}\\
&& $20\%$& {\bf 0.8269}& 0.7111& 0.8087& 0.8205& {\bf 0.8941}& 0.7859& {\bf 0.8378}\\
&& $40\%$& {\bf 0.8674}& 0.7375& 0.8104& 0.8347& {\bf 0.9153}& 0.8167& {\bf 0.8530}\\
\cline{2-10}
&\multirow{4}{*}{$40\%$}
& $5\%$& 0.7379& 0.7182& {\bf 0.7789}& {\bf 0.8164}& 0.8123& 0.7396& {\bf 0.7998}\\
&& $10\%$& 0.7730& 0.7353& {\bf 0.7954}& 0.8370& {\bf 0.8561}& 0.7351& {\bf 0.8210}\\
&& $20\%$& {\bf 0.8332}& 0.7817& 0.8087& 0.8392& {\bf 0.8941}& 0.7813& {\bf 0.8378}\\
&& $40\%$& {\bf 0.8724}& 0.8097& 0.8104& 0.8639& {\bf 0.9153}& 0.8038& {\bf 0.8530}\\
\cline{2-10}
&\multirow{4}{*}{$60\%$}
& $5\%$& 0.7376& 0.7445& {\bf 0.7789}& {\bf 0.8132}& 0.8123& {\bf 0.8051}& 0.7998\\
&& $10\%$& 0.7778& 0.7831& {\bf 0.7954}& 0.7639& {\bf 0.8561}& 0.7444& {\bf 0.8210}\\
&& $20\%$& {\bf 0.8367}& 0.8264& 0.8087& 0.8251& {\bf 0.8941}& 0.7755& {\bf 0.8378}\\
&& $40\%$& {\bf 0.8753}& 0.8504& 0.8104& 0.8716& {\bf 0.9153}& 0.7899& {\bf 0.8530}\\
\hline \hline
\multirow{12}{*}{autofood}&\multirow{4}{*}{$20\%$}
& $5\%$& {\bf 0.7170}& 0.5198& 0.6451& {\bf 0.7070}& 0.6356& {\bf 0.7235}& 0.6450\\
&& $10\%$& {\bf 0.8083}& 0.5578& 0.7576& {\bf 0.8194}& 0.7259& {\bf 0.8131}& 0.7576\\
&& $20\%$& 0.8043& 0.5804& {\bf 0.8178}& {\bf 0.8797}& 0.7712& {\bf 0.8665}& 0.8178\\
&& $40\%$& 0.8007& 0.5807& {\bf 0.8860}& {\bf 0.9317}& 0.8087& {\bf 0.9237}& 0.8857\\
\cline{2-10}
&\multirow{4}{*}{$40\%$}
& $5\%$& {\bf 0.7129}& 0.6299& 0.6451& {\bf 0.7029}& 0.6356& {\bf 0.7157}& 0.6450\\
&& $10\%$& {\bf 0.8218}& 0.6517& 0.7576& {\bf 0.8198}& 0.7259& {\bf 0.8175}& 0.7576\\
&& $20\%$& {\bf 0.8634}& 0.6322& 0.8178& {\bf 0.8796}& 0.7712& {\bf 0.8644}& 0.8178\\
&& $40\%$& 0.8131& 0.6848& {\bf 0.8860}& {\bf 0.9319}& 0.8087& {\bf 0.9260}& 0.8857\\
\cline{2-10}
&\multirow{4}{*}{$60\%$}
& $5\%$& {\bf 0.7175}& 0.6013& 0.6451& {\bf 0.7045}& 0.6356& {\bf 0.7128}& 0.6450\\
&& $10\%$& {\bf 0.8206}& 0.6316& 0.7576& {\bf 0.8196}& 0.7259& {\bf 0.8213}& 0.7576\\
&& $20\%$& {\bf 0.8725}& 0.6758& 0.8178& {\bf 0.8800}& 0.7712& {\bf 0.8781}& 0.8178\\
&& $40\%$& 0.8141& 0.6351& {\bf 0.8860}& {\bf 0.9315}& 0.8087& {\bf 0.9255}& 0.8857\\
\hline \hline
\multirow{12}{*}{compphys}&\multirow{4}{*}{$20\%$}
& $5\%$& {\bf 0.6486}& 0.5727& 0.6457& {\bf 0.6479}& 0.6424& {\bf 0.6488}& 0.6457\\
&& $10\%$& {\bf 0.7478}& 0.5691& 0.7235& {\bf 0.7473}& 0.7147& {\bf 0.7556}& 0.7235\\
&& $20\%$& {\bf 0.7908}& 0.5729& 0.7459& {\bf 0.7921}& 0.7297& {\bf 0.8101}& 0.7459\\
&& $40\%$& {\bf 0.8172}& 0.6788& 0.7728& {\bf 0.8416}& 0.7413& {\bf 0.8718}& 0.7730\\
\cline{2-10}
&\multirow{4}{*}{$40\%$}
& $5\%$& {\bf 0.6474}& 0.6049& 0.6457& {\bf 0.6478}& 0.6424& {\bf 0.6480}& 0.6457\\
&& $10\%$& {\bf 0.7509}& 0.6295& 0.7235& {\bf 0.7481}& 0.7147& {\bf 0.7437}& 0.7235\\
&& $20\%$& {\bf 0.7964}& 0.6442& 0.7459& {\bf 0.7913}& 0.7297& {\bf 0.7849}& 0.7459\\
&& $40\%$& {\bf 0.8192}& 0.6651& 0.7728& {\bf 0.8371}& 0.7413& {\bf 0.8561}& 0.7730\\
\cline{2-10}
&\multirow{4}{*}{$60\%$}
& $5\%$& 0.6443& 0.6089& {\bf 0.6457}& {\bf 0.6468}& 0.6424& {\bf 0.6601}& 0.6457\\
&& $10\%$& {\bf 0.7504}& 0.6505& 0.7235& {\bf 0.7489}& 0.7147& {\bf 0.7421}& 0.7235\\
&& $20\%$& {\bf 0.7991}& 0.6687& 0.7459& {\bf 0.7854}& 0.7297& {\bf 0.8064}& 0.7459\\
&& $40\%$& {\bf 0.8269}& 0.7240& 0.7728& {\bf 0.8378}& 0.7413& {\bf 0.8659}& 0.7730\\
\end{tabular}
}
\end{table}

\end{document}